\documentclass[11pt]{article}

\usepackage{amsmath, amssymb}
\usepackage{graphicx}
\usepackage{hyperref}
\usepackage{natbib}          
\usepackage{geometry}        
\usepackage{times}           
\usepackage{float}           
\usepackage{booktabs}        
\usepackage{caption}
\usepackage{subcaption}
\usepackage{svg}
\usepackage{color}
\usepackage{booktabs}
\usepackage{multirow}
\usepackage[normalem]{ulem}
\useunder{\uline}{\ul}{}
\usepackage{listings}

\lstdefinestyle{llmprompt}{
    backgroundcolor=\color{gray!10},
    basicstyle=\ttfamily\small,
    frame=single,
    breaklines=true,
    columns=fullflexible,
    keepspaces=true,
    captionpos=b,
    showstringspaces=false
}

\lstdefinestyle{caption}{
    backgroundcolor=\color{gray!5},
    basicstyle=\ttfamily\tiny,
    breaklines=true,
    columns=fullflexible,
    keepspaces=true,
    captionpos=b,
    showstringspaces=false
}

\title{VIVECaption: A Split Approach to Caption Quality Improvement}
\author{
  Varun Ananth\thanks{Corresponding author} \\
  \texttt{vananth@adobe.com} \\
  \normalsize Adobe Inc. \\
  \and
  Baqiao Liu \\
  \normalsize \texttt{baqiaol@adobe.com} \\
  \normalsize Adobe Inc. \\
  \and
  Haoran Cai \\
  \normalsize \texttt{hcai@adobe.com} \\
  \normalsize Adobe Inc. \\
}
\date{}

\begin{document}

\maketitle

\begin{abstract}
Caption quality has emerged as a critical bottleneck in training high-quality text-to-image (T2I) and text-to-video (T2V) generative models. While visual language models (VLMs) are commonly deployed to generate captions from visual data, they suffer from hallucinations, poor compositional reasoning, and limited fine-grained understanding, resulting in misaligned image-caption pairs that degrade downstream model performance. This technical report introduces VIVECaption, a systematic two-sided approach to caption quality improvement. We first establish a comprehensive taxonomy of caption evaluation metrics, distinguishing between "universal" and "instance-grounded" metrics, with the ultimate goal of showcasing the use-cases and tradeoffs between different caption quality metrics. We then use this language to describe our two-sided approach to caption quality improvement: (1) a gold-standard dataset creation methodology using stratified sampling and (2) a model alignment strategy encompassing context alignment and parameter-level finetuning using SFT. We demonstrate our methodology on open-source models, focusing on structured caption formats that enable better parsing and downstream utilization. We ultimately show that using a finetuned character detection model in an image captioning pipeline significantly improves holistic image-caption alignment quality. Our work addresses the growing need for high-quality "vegan" training data in enterprise AI development, providing practical solutions for teams seeking to improve caption-image alignment without relying on potentially copyright-protected web-scraped content.
\end{abstract}





\section{Introduction}

Diffusion models have become a cornerstone of modern generative AI, demonstrating state-of-the-art performance in domains like image, video, and audio synthesis following extensive pre-training~\citep{ho2020denoising}. However, a significant limitation of these foundational models is their struggle to generalize to novel concepts not well-represented in their initial training data. This generalization gap has spurred a widespread shift towards model personalization and finetuning, where practitioners adapt pre-trained models to specialized domains, styles, or subjects~\citep{ruiz2022dreambooth}. The result has been a proliferation of highly specialized models, a trend evidenced by the rapid growth of community platforms like CivitAI, which host vast collections of models finetuned for specific purposes.

During practical finetuning, due to the large-scale existence of images, often the practical bottleneck is the caption quality. Visual language models~\citep{radford2021learning}, despite recent advancements, remain imperfect and struggle with several key limitations. Notably, they exhibit poor compositional reasoning, failing to correctly describe the relationships between objects in a scene. They also have limited fine-grained understanding, often missing subtle but important details like text within an image. Finally, they are prone to hallucinating objects based on statistical priors from their training data rather than the visual evidence itself, further degrading caption accuracy~\citep{li2023grounded}.

\begin{figure}[H]
    \centering
    \includegraphics[width=0.8\textwidth]{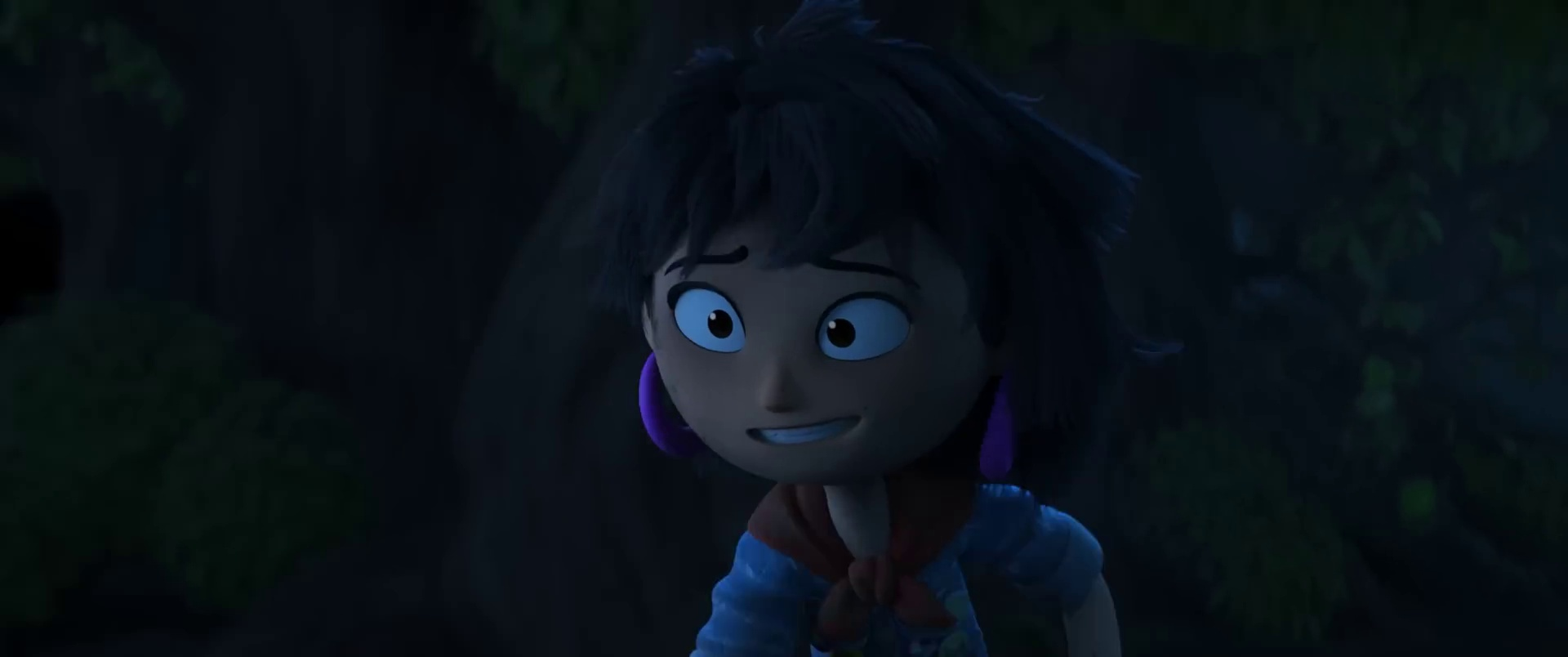}
    \begin{lstlisting}[style=caption]
        {
            "scene": "The image shows a character named Victoria in a dark, forest-like setting. She appears to be looking forward with a concerned expression.",
            "background": "The background is dark and shadowy, with hints of foliage, suggesting a nighttime forest or cave setting.",
            "characters": {
                "Victoria": {
                    "description": "Victoria has dark hair, large expressive eyes, and is wearing a blue shirt with a red scarf and purple earrings.",
                    "location": "Center of the image",
                    "expression": "Concerned",
                    "pose": "Standing and looking forward"
                }
            },
            "salient_objects": {}
        }
    \end{lstlisting}
    \caption{\centering As seen in [\ref{fig:character_grid}], the character in the image above is Ellie not Victoria. However, the caption claims Victoria to be the focus of the image. This datapoint, if used to train a downstream T2I model, would actively harm the model's performance.}
    \label{fig:misaligned_img_cap}
\end{figure}

Data quality is increasingly becoming the core focus of GenAI engineering teams, taking precedence over traditionally important features like model architecture or computational techniques. This emphasis on data is made more critical if a team wishes to only use ``vegan'' data, without scraping the internet for possibly protected intellectual property. A common problem teams face when creating a T2I or T2V model is that the data, namely (image, caption) or (video, caption) pairs, are not well harmonized. VLMs are often deployed to create the captions based on the image, and are prone to hallucinations. This ultimately has downstream effects, lowering the quality of the delivered T2I or T2V model. To illustrate, [\ref{fig:misaligned_img_cap}] is an example of a misaligned image-caption pair.

The purpose of this paper is to show off an easy yet effective way to improve image-caption alignment holistically using supervised finetuning (SFT) with open-source VLMs. A core component of this method is the creation of a golden-standard dataset, a step we urge teams to take seriously. We also introduce a metric taxonomy so that teams can tailor our general purpose method to their goals.

\section{Background}
\subsection{Modes of Captioning}
Image captions can be broadly categorized into two types: \textbf{structured} and \textbf{dense} (or unstructured). Structured captions offer clarity and consistency through a predefined format, making them ideal for systematic data extraction. An example of a structured caption would be one where different aspects of a caption (i.e. scene, background, characters, mood) would each be a key within a dictionary, with the values describing that aspect of the image. Dense captions often contain these elements as well, but mix them all together using natural language. The result is more of a descriptive paragraph. Structured captions provide certain guarantees of the content within a caption, and the atomization of the caption can make evaluation easier. Dense captions provide creative flexibility and emotional engagement, but they can be ambiguous.  The choice of captioning mode thus involves a trade-off between structural rigidity, and narrative flexibility.

We decided that structured captions provide the better half of this tradeoff, partially due to the fact that structured captions can more easily be collapsed into dense captions than the other way around. Furthermore, evidence suggests that providing a strict template for an LLM to follow improves its reasoning capabilities ~\citep{awal2025investigatingpromptingtechniqueszero}. Henceforth, all captions can be assumed to have some level of structure to them.

\subsection{Taxonomy of Caption Metrics}
\begin{figure}[H]
    \centering
    \includegraphics[width=0.8\textwidth]{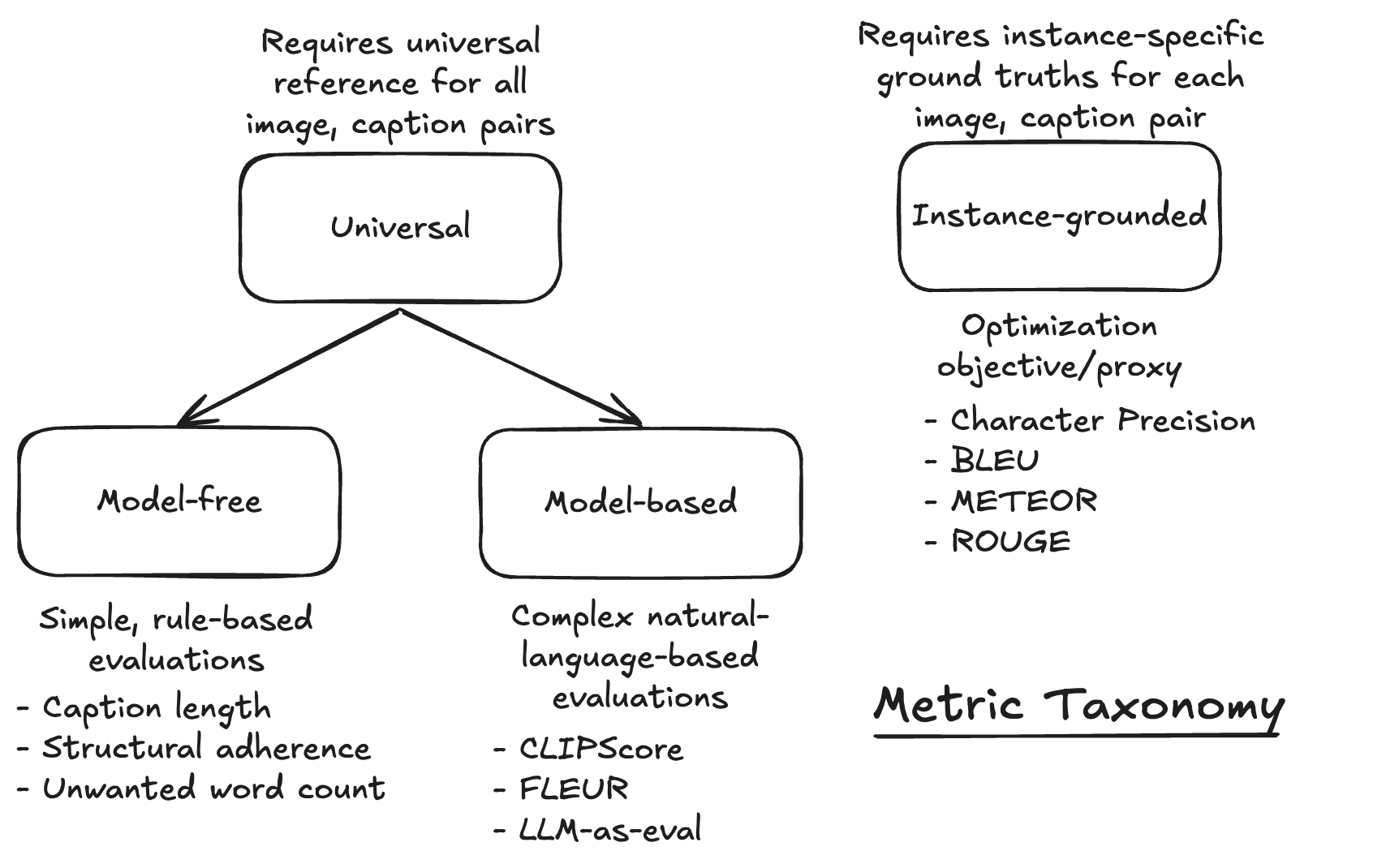}
    \caption{\centering Diagram illustrating the large categories of caption quality metrics: universal and instance-grounded. The two sub-categories of universal metrics are also displayed, with examples for all types.}
    \label{fig:metric_taxonomy}
\end{figure}

In order to improve the alignment between an image, caption pair from a dataset $((i,c) \in \mathcal{D})$, there must be an axis on which to evaluate their alignment. Metrics provide this marker and it is therefore worthwhile to taxonomize the different types of calculable metrics, and their tradeoffs. The output of all of these metric types is some score $s$, which could be an ordinal, discrete, or continuous variable but not nominal as some level of ranking is necessary. A metric $\mathcal{M}$ can be defined as a mapping: $\mathcal{M}: (i,c) \rightarrow s$

A \textbf{universal} metric is one that is directly calculable from the $(i,c)$ pair and some universal reference $\mathcal{R}$. $\mathcal{R}$ can be some decided truth relevant to the dataset as a whole (\textbf{model-free universal}) or embedded within another model's weights (\textbf{model-based universal}). For the former an example is caption length with $\geq$ 1024 tokens being considered ``too long'', and for the latter an example is CLIP score ~\citep{radford2021learning}. An \textbf{instance-grounded} metric is one that requires some additional reference $r$ for \textit{each} (image, caption) pair during calculation. There is no important distinction here between $r$ generated by a model or not. An example of an instance-grounded metric would be character precision, requiring a set of ground-truth characters per $(i,c)$ pair. Another example would be using a very large model to generate a target caption $r$ from $i$, and then computing the BLEU score ~\citep{papineni2002bleu} between $c$ and $r$.

Each of these types of caption quality metrics have different niches, and can be put together for a holistic picture of data quality. Model-free universal metrics like caption length or structured output adherence cannot capture natural language semantics, limiting their usefulness in improving holistic caption quality but shine in their ease-of-use. It is easy to scan a large dataset and aggregate values for universal model-free metrics. These scans can help root out obvious data quality issues early on. Universal model-based metrics \textit{can} capture natural language semantics and are very flexible in what they evaluate. On the other hand, the utility of these metrics as optimization objectives is limited. If a universal metric immediately maps $(i,c) \rightarrow s$, only RL-based finetuning methods are available to improve caption quality if the score is interpreted as a reward. Self-supervised training may also be available as a way to improve caption quality.

Universal metrics do not require a hand-annotated as the universal reference $\mathcal{R}$ can be used across the entire dataset without the need for any kind of ``label'' per image-caption pair. Instance-grounded metrics are the best as an optimization objective because instead of being limited to RL-based VLM finetuning methods, traditional SFT becomes available. The downside to instance-grounded metrics is that they require manual curation of a golden-standard dataset, limiting the data coverage of their evaluation. As we show below, however, the benefits of curating even a small dataset in a principled way can have a significant impact on caption quality. In general: model-free universal metrics are useful for \textbf{simple rule-based evaluation}, model-based universal metrics are useful for \textbf{complex natural-language-based evaluation}, and instance-grounded metrics are more useful as an \textbf{optimization objective}.

\subsection{Suggestions for Other Teams}\label{sec:suggestions}
The above metric taxonomy will hopefully prove useful for teams looking for a shared language to describe the different axes on which captions can be improved. We generally suggest teams create a set of metrics that they feel covers their bases, and then use them as follows:
\begin{itemize}
    \item \textbf{Model-free universal:} Elucidate simple information about the entire dataset, \textit{initial ``health check'' of the data}
    \item \textbf{Instance-grounded}: Create a gold-standard dataset with labels $r$ for each $(i,c)$, and use your finetuning method of choice to improve your captioning model \textit{with improving this metric as the optimization objective}
    \item \textbf{Model-based universal:} Holistic evaluation used as a way to \textit{compare $(i,c)$ pairs before and after improvement} without the need for additional hand-labeled data.
\end{itemize}

\section{Caption Quality Improvement Experiment}
We aim to demonstrate the efficacy of our two-sided approach VIVECaption with an experiment on entirely open-source data. ``Sprite Fright'' is an open-source blender animated short movie, with nearly all assets open to the public. It includes 7 main characters throughout the film, with varying prominence. With the ultimate goal of training a theoretical T2I model (future work), we use the VIVECaption approach to significantly improve the alignment between the $(i,c)$ pairs gathered by sampling frames from the movie. We also demonstrate the utility of using the metric taxonomy as described in [\ref{sec:suggestions}].
\begin{figure}[H]
    \centering
    \includegraphics[width=0.7\textwidth]{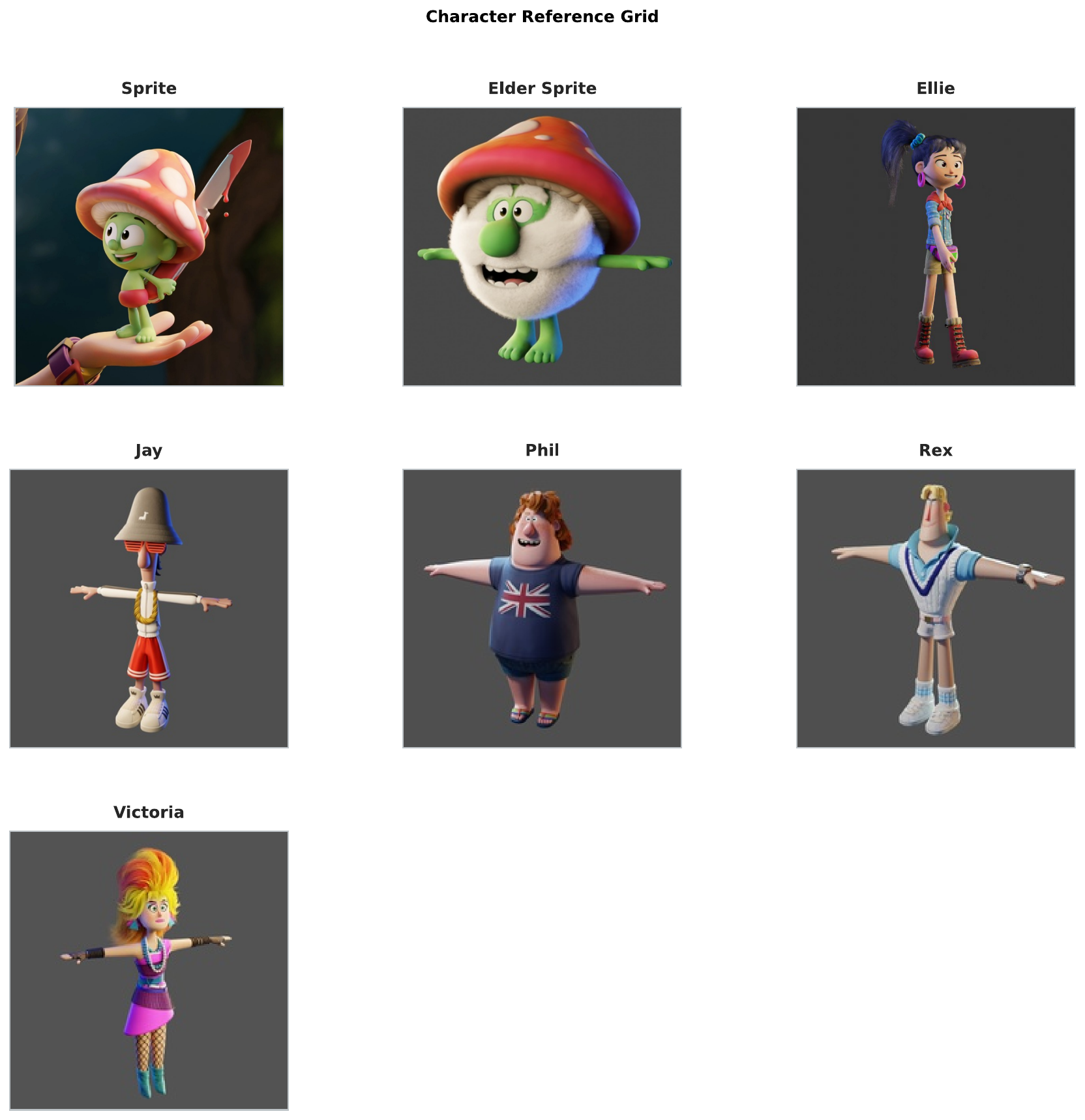}
    \caption{\centering Characters from ``Sprite Fright''. These images were used for in-context alignment as a reference to both the character detection model and image captioning model. Providing images allows the VLM to ``understand'' the language particular to this short film. Although obviously preferable, the images do not need to be ``uniform'' in character posture or angle.}
    \label{fig:character_grid}
\end{figure}
The axis of improvement we have chosen to pursue in this study broadly falls into the idea of ``character consistency''. For many media settings, characters interactions comprise a great amount of the describable action within a scene. As such, ensuring that the characters \textit{claimed} to be in the image by the captioning model actually \textit{are} in the image should improve caption quality overall. Keeping in the spirit of ``vegan'' data and models, we only use open source model classes, limiting ourselves to Qwen2.5-VL ~\citep{bai2025qwen25vltechnicalreport} and InternVL3 ~\citep{zhu2025internvl3exploringadvancedtraining} as possible captioning models. Using these models we define a multi-step captioning pipeline, with Qwen2.5-VL (3B, 7B, or 32B) being used to detect characters and passing this information to InternVL3-38B to generate a structured caption (see [\ref{appendix:vlm_prompts}]). By aligning the character detection Qwen2.5-VL model on instance-grounded metrics, we hope to improve the overall quality of the captions. 

Given a set $r$ of characters that are in an image $i$ (from a ground-truth dataset) and a prediction set $\hat{r}$ from Qwen2.5-VL, we can define MacroF1, Precision, Recall and \# Mistakes as instance-grounded metrics.

$$\textrm{MacroF1} = \frac{2 \times \mathrm{Precision} \times \mathrm{Recall}}{\mathrm{Precision} + \mathrm{Recall}}$$
$$\mathrm{Precision} = \frac{TP}{TP + FP}$$
$$\mathrm{Recall} = \frac{TP}{TP + FN}$$
$$\mathrm{\#\;Mistakes} = FP + FN$$
Where:
$$TP = |r \cap \hat{r}|,\;FP = |r \setminus \hat{r}|,\; FN = |\hat{r} \setminus r|$$

Averaged over all $|\mathcal{D}|$ examples, we get an idea of how consistently the characters are being identified in our captioning pipeline.

At the end of the alignment process, we can verify that the captions truly have gotten better by using a SOTA model like Gemini-2.5-Pro to score the captions with model-based universal metrics. Our captioning pipeline provides us $(i,c)$ pairs where $c$ is a dictionary containing ``scene'', ``background'', ``characters'', and ``salient objects''. Each one of these can be evaluated numerically from 1-10 based on a comprehensive rubric provided to the evaluation model (see [\ref{appendix:vlm_prompts}]). To perform any of these experiments, we must first intelligently construct a gold-standard dataset. We outline this process below.

\section{Side A: Gold-Standard Dataset}
\subsection{Sampling Techniques and Motivations}
As mentioned previously, creating a gold-standard dataset is a prerequisite to aligning on instance-grounded metric. As we are analyzing character consistency, a near-even distribution of all relevant characters from the dataset would be ideal. However, acquiring this becomes a chicken-and-egg problem. If we have a VLM that can detect the characters very consistently, allowing us to perform stratified sampling after running a detection pass, then there is no need for alignment. Vice versa, if we cannot easily identify characters it becomes hard to construct a stratified sample as the identifications may be hallucinated. 

\begin{figure}[H]
    \centering
    \begin{table}[H]
    \centering
    \begin{tabular}{cc}
        \includegraphics[width=0.4\textwidth]{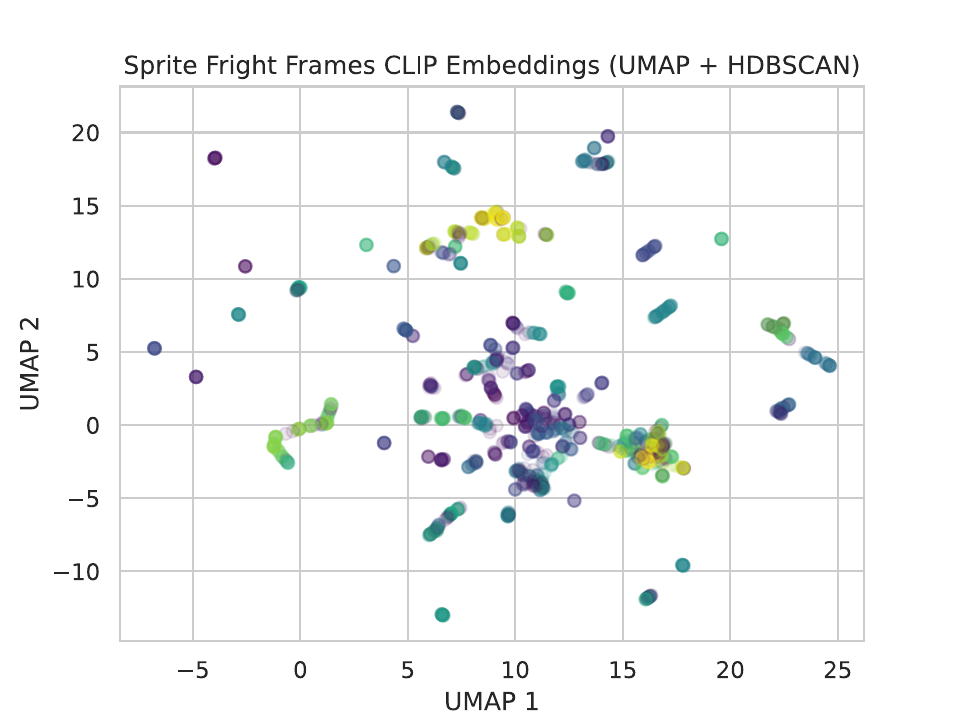} &
        \includegraphics[width=0.4\textwidth]{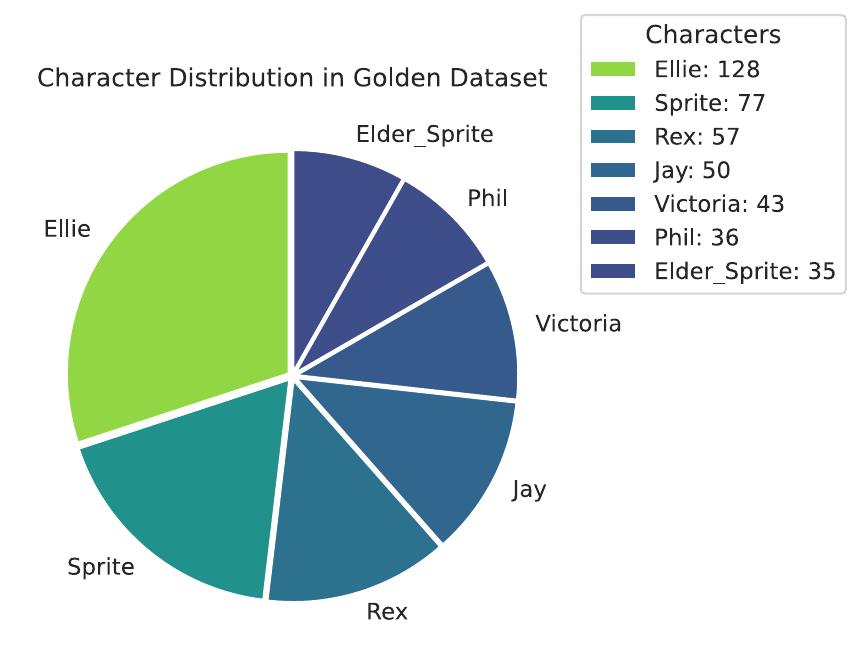}
    \end{tabular}
    \end{table}
    \caption{\centering (Left) UMAP projection onto two dimensions of CLIP embeddings of all sampled frames. Color distinguishes each of the 310 ``clusters'' as located by HDBSCAN. Samples from each cluster were used to make a gold-standard dataset. (Right) Pie chart showing the occurrence of each character in the gold-standard dataset after stratified sampling on HDBSCAN clusters. Ellie's over-representation is a consequence of her being the main character and is essentially unavoidable. The total number of samples in the gold standard dataset is 310, which is 14.35\% of the entire 2161 frame dataset.}
    \label{fig:clip_embeds}
\end{figure}

We opted to circumvent this by instead clustering image embedding vectors and sampling evenly from those clusters. We start with our ``full'' dataset, which consists of 2161 frames at 1080p, sampled in $\frac{1}{4}$ second intervals for the duration of the short film. From this, we used CLIP ViT-B/32 to embed all the images into 512-length vectors. [\ref{fig:clip_embeds}] shows these embeddings projected into 2D space with UMAP, and then clustered into 310 clusters with HDBSCAN.

We then sampled one image from each of these clusters to generate our golden-standard dataset. After the human annotation procedure with LabelStudio ~\citep{labelstudio}, we plot the distribution of characters in our dataset. While not perfectly even, this distribution makes sense as ``Ellie'' is the main character of the short film and will therefore be over-represented. The representation of this character seems to be within reasonable bounds. We advise any teams following this procedure to ensure that the character distribution of their golden-standard dataset is to their liking. We then labeled each of the 310 images with the characters present in the image. If a replicated character (i.e. ``Sprite'') appears in a scene, we only tag it once. Each image now has a set of ground-truth characters $r$ associated with it. We can now perform SFT on Qwen2.5-VL models with $r$ as target VLM output.

\section{Side B: Model Alignment}
\subsection{Context Alignment}
In-context alignment is the easiest form of alignment to perform as it requires no parameter updates. LLMs and VLMs have been shown to ``learn'' from context primers, efficiently completing tasks like labeling, item detection, pattern matching, etc ~\citep{brown2020fewshot}. For our case, we perform in-context alignment with the Qwen family of VLMs by interleaving images of target characters with their descriptions, and capping off the context with detailed instructions on how to process the provided image (see [\ref{appendix:vlm_prompts}]). The quality of captions after a character detection step with an off-the-shelf model with in-context alignment becomes our effective baseline. As a further control, we also include in our analyses an even more basic captioning pipeline: just providing all characters as context and requesting a VLM to caption the image. As we show, aligning the character detection model on the aforementioned gold-standard dataset significantly improves caption quality.

\subsection{Parameter Alignment}
We outline here the general process used to align our character detection VLMs. Basic SFT was our finetuning method of choice as we found it was effective yet incredibly simple to implement. We created an 80/20 train/test split by randomly sampling from our gold standard dataset. We did not do full parameter finetuning and opted for parameter efficient fine tuning (PeFT) instead. We found empirically that this improved results, but a more thorough examination is one we leave to future work. Only 5 epochs of training were needed with a learning rate of 1e-4 for the train and validation losses to plateau (see [\ref{appendix:train_curves}]). In order to have the target labels for images be consistent, the character set was alphabetically ordered when made the target output of the model. The models were prompted to output a list of character names that are in a target image, given context about the characters that could possibly be in the image. This simple alignment procedure drastically improved caption quality when the finetuned VLM was used as a character detector in the captioning pipeline.

\section{Results}
\subsection{Character Detection Model Improvements}
\begin{figure}[H]
    \centering
    \begin{table}[H]
    \centering
    \begin{tabular}{cc}
        \includegraphics[width=0.4\textwidth]{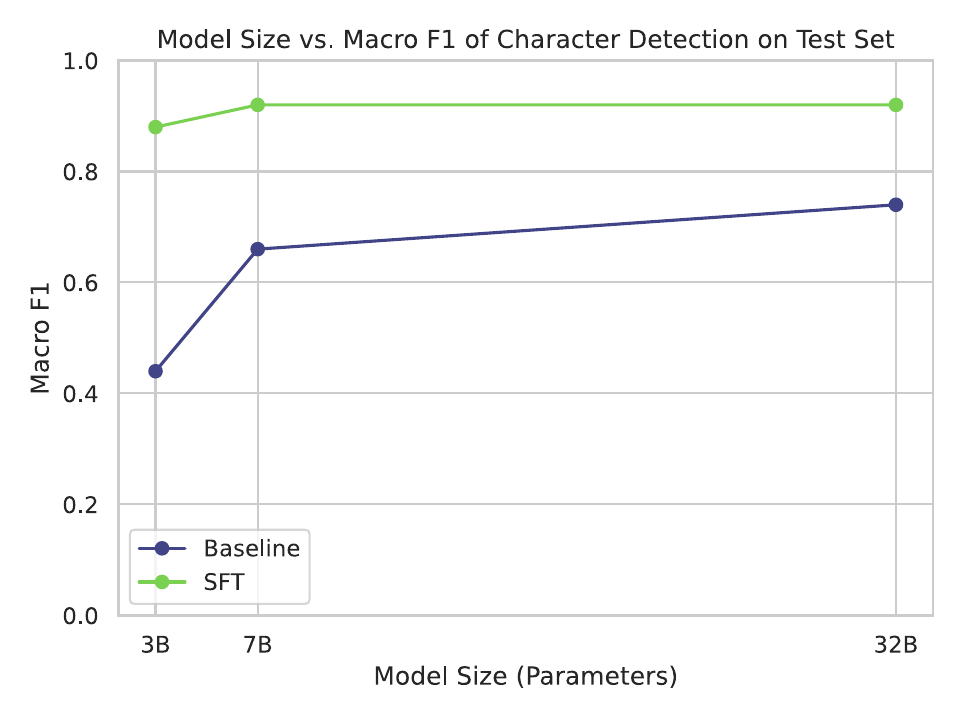} &
        \includegraphics[width=0.4\textwidth]{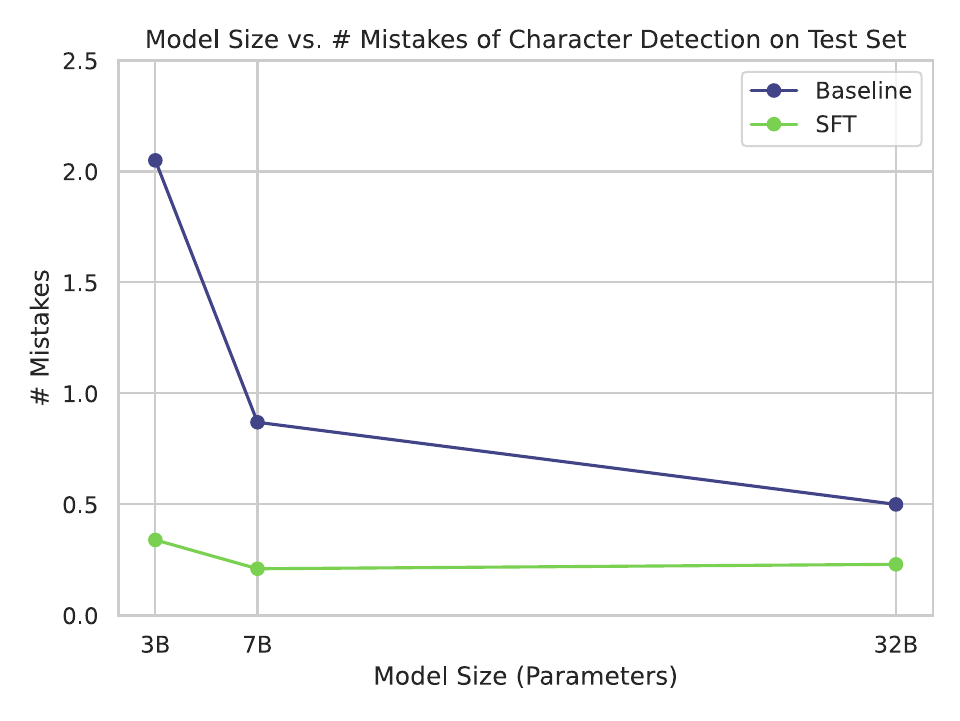}
    \end{tabular}
    \end{table}
    \caption{\centering (Left) Plot showing relative Qwen2.5-VL parameter sizes and the effect on the average Macro F1 score on the test set pre-and-post SFT. The greatest jump in performance for both the baseline and SFT comes between the 3B and 7B parameter models. (Right) Plot showing relative Qwen2.5-VL parameter sizes and the effect on the average \# of mistakes (per example) on the test set pre-and-post SFT. The greatest jump in performance for both the baseline and SFT comes between the 3B and 7B parameter models.}
    \label{fig:f1_mistakes_plot}
\end{figure}

As shown in [\ref{fig:f1_mistakes_plot}], the finetuned models greatly outperform the off-the-shelf ones across the board as expected. Perhaps surprising is that the significant jump in quality between the off-the-shelf 3B and 7B models was essentially removed by the finetuning process. The finetuned 3B and 7B models perform comparably on the MacroF1 and \# mistakes metrics.

The difference in quality between 7B and 32B models is negligible for both the off-the-shelf and finetuned versions, possibly indicating that for this task an increase in parameters does not correlate linearly with model quality. A table summary of this information can be found in [\ref{tab:main_metric_results}]. The finetuned versions of the models always outperform their non-finetuned counterparts. These improvements on instance-grounded metrics translate to improvements in universal model-based metrics as we show in the following section.

\begin{table}[H]
\centering
\begin{tabular}{l|cccccc}
\hline
\multicolumn{1}{c|}{\textbf{Metric}} & \multicolumn{6}{c}{\textbf{Model Parameters}}                                                                                                       \\ \hline
\multirow{2}{*}{}                    & \multicolumn{2}{c}{{\ul 3B}}                  & \multicolumn{2}{c}{{\ul 7B}}                           & \multicolumn{2}{c}{{\ul 32B}}     \\ \cline{2-7} 
                                     & \textit{Baseline} & \textit{SFT}              & \textit{Baseline} & \textit{SFT}                       & \textit{Baseline} & \textit{SFT}  \\ \cline{2-7} 
\textbf{Precision $\uparrow$}  & 0.46    & \multicolumn{1}{c|}{\textit{0.91}} & 0.71  & \multicolumn{1}{c|}{\textbf{\textit{0.92}}} & 0.75   & \textit{0.91} \\
\textbf{Recall $\uparrow$}   & 0.52   & \multicolumn{1}{c|}{\textit{0.87}} & 0.64   & \multicolumn{1}{c|}{\textit{0.93}}   & 0.75     & \textbf{\textit{0.94}} \\
\textbf{Macro F1 $\uparrow$}  & 0.44    & \multicolumn{1}{c|}{\textit{0.88}} & 0.66  & \multicolumn{1}{c|}{\textbf{\textit{0.92}}} & 0.74   & \textbf{\textit{0.92}} \\
\textbf{\# Mistakes $\downarrow$}   & 2.05   & \multicolumn{1}{c|}{\textit{0.34}} & 0.87   & \multicolumn{1}{c|}{\textbf{\textit{0.21}}} & 0.50  & \textit{0.23}  \\ \hline
\end{tabular}
\caption{\centering Table showing the values on the test set for the following instance-grounded metrics: precision, recall, MacroF1, and \# of mistakes. In bold are the best values for each metric across all model sizes and baseline vs. SFT. Italicized values indicate the best performance on a metric within a model size. SFT always outperforms the baseline.}
\label{tab:main_metric_results}
\end{table}

\subsection{Holistic Caption Quality Evaluation}
\begin{figure}[H]
    \centering
    \begin{table}[H]
    \centering
    \begin{tabular}{cc}
        \includegraphics[width=0.4\textwidth]{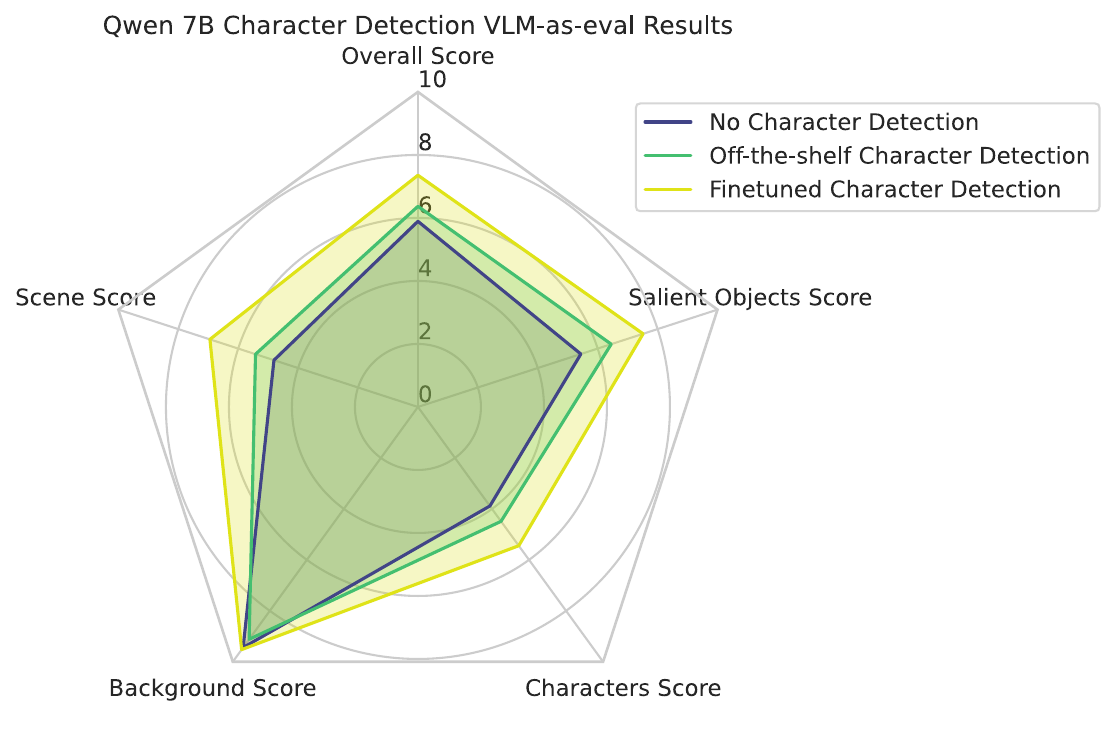} &
        \includegraphics[width=0.4\textwidth]{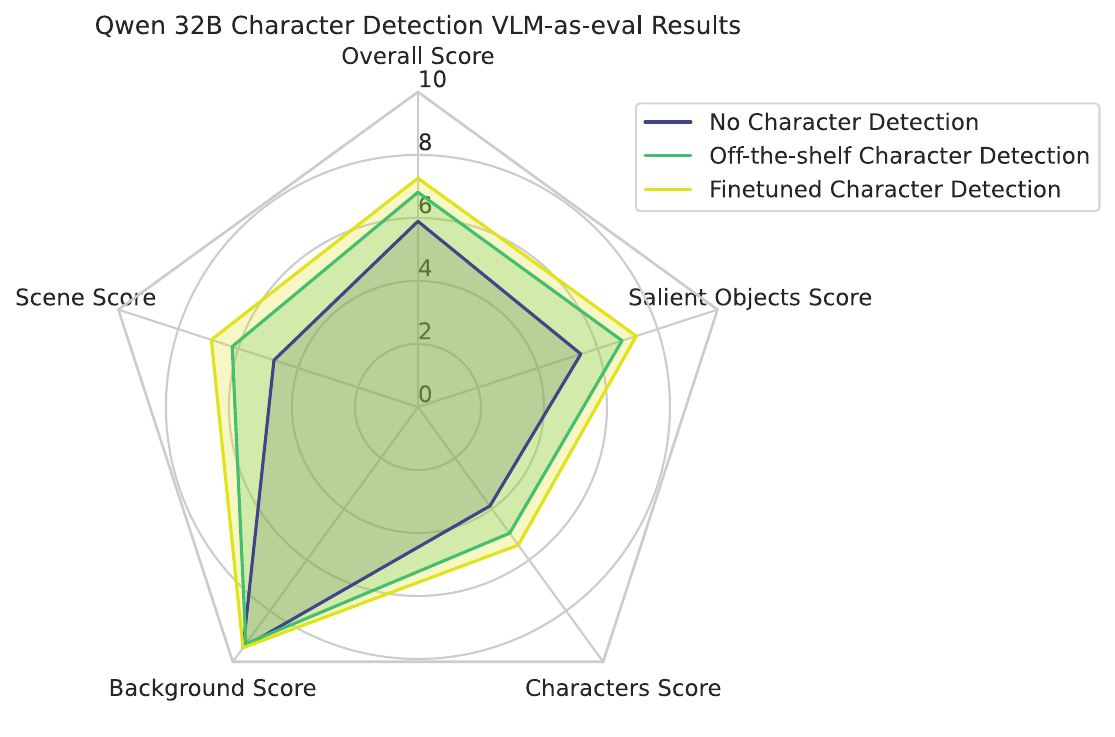}
    \end{tabular}
    \end{table}
\caption{\centering Spider charts showing the VLM-as-eval metric improvements for the two-stage captioning pipeline when a Qwen2.5-VL 7B or 32B model is employed for character detection. Also shown are the evaluation scores for single-pass captioning without a character detection stage and all characters passed as context. 300 random images not in the train/test set for finetuning were selected to be captioned and evaluated. The caption quality improvements when the model is fine tuned with SFT carry beyond just the ``character'' section of the structured caption.}
\label{fig:llm_as_eval_charts}
\end{figure}

We now move to see if our improved captioning pipeline not only leads to an increase in instance-grounded metrics across the gold-standard dataset, but an increase in caption quality when evaluated on many axes by a VLM. We chose four: salient objects, characters, background, and scene. Overall score is the arithmetic mean of these four categories. The prompt provided to Gemini-Pro-2.5 to evaluate image-caption pairs can be found in [\ref{appendix:vlm_prompts}]. Seeing as the 3B model consistently performed worse on the previous metrics, we restricted our focus to 7B and 32B models. As seen numerically in [\ref{tab:llm_as_eval_metric_results}] and graphically in [\ref{fig:llm_as_eval_charts}], using a fine-tuned character detection model improved caption quality across the board. The images we evaluate on here are 300 randomly sampled from all frames not in the gold-standard dataset, meaning that there is no chance for train/test leakage here and that these improvements are ``real''. To further show that these improvements are not due to chance within the inherently stochastic VLM-as-eval paradigm, we perform a paired t-test across image-caption pairs generated without a finetuned character detection model and with one. As shown in [\ref{tab:llm_as_eval_pvalue}], we receive exceptionally small p-values for all scores except ``background score''.

\begin{table}[H]
\begin{tabular}{l|ccccc}
\hline
\multicolumn{1}{c|}{\textbf{Metric}} & \multicolumn{5}{c}{\textbf{Model Parameters}}                                                                                                           \\ \hline
\multirow{2}{*}{}                    & \multicolumn{1}{c|}{{\ul No Character Detection}} & \multicolumn{2}{c|}{{\ul 7B}}                                   & \multicolumn{2}{c}{{\ul 32B}}     \\ \cline{2-6} 
                                     & \multicolumn{1}{c|}{\textit{Baseline}}            & \textit{Baseline} & \multicolumn{1}{c|}{\textit{SFT}}           & \textit{Baseline} & \textit{SFT}  \\ \cline{2-6} 
\textbf{Overall Score} (/10)                  & \multicolumn{1}{c|}{5.89}                         & 6.36              & \multicolumn{1}{c|}{\textit{\textbf{7.35}}} & 6.82              & \textit{7.26} \\
\textbf{Salient Objects Score} (/10)          & \multicolumn{1}{c|}{5.44}                         & 6.45              & \multicolumn{1}{c|}{\textit{\textbf{7.52}}} & 6.81              & \textit{7.29} \\
\textbf{Characters Score} (/10)               & \multicolumn{1}{c|}{3.89}                         & 4.48              & \multicolumn{1}{c|}{\textit{\textbf{5.44}}} & 4.95              & \textit{5.42} \\
\textbf{Background Score} (/10)               & \multicolumn{1}{c|}{9.44}                         & 9.11              & \multicolumn{1}{c|}{\textit{\textbf{9.52}}} & 9.29              & \textit{9.45} \\
\textbf{Scene Score} (/10)                    & \multicolumn{1}{c|}{4.81}                         & 5.42              & \multicolumn{1}{c|}{\textit{\textbf{6.94}}} & 6.20              & \textit{6.90} \\ \hline
\end{tabular}
\caption{\centering Table showing the values for the following VLM-as-eval metrics on 300 randomly sampled images not in the train/test sets: salient objects, characters, background, and scene. Overall score is a simple mean of all the aforementioned values. In bold are the best values for each metric across all model sizes and baseline vs. SFT. Italicized values indicate the best performance on a metric within a model size. SFT always outperforms the baseline.}
\label{tab:llm_as_eval_metric_results}
\end{table}

\begin{table}[H]
\centering
\begin{tabular}{l|ll}
\hline
\multicolumn{1}{c|}{\textbf{Metric}} & \multicolumn{2}{c}{\textbf{Model Parameters}}                 \\ \hline
\multirow{2}{*}{}                    & \multicolumn{1}{c|}{{\ul 7B}} & \multicolumn{1}{c}{{\ul 32B}} \\ \cline{2-3} 
                                     & \multicolumn{2}{c}{\textit{p-value}}                          \\ \cline{2-3} 
\textbf{Overall Score}               & \multicolumn{1}{l|}{6.72e-15} & 4.48e-05                       \\
\textbf{Salient Objects Score}       & \multicolumn{1}{l|}{1.82e-10} & 1.25e-03                         \\
\textbf{Characters Score}            & \multicolumn{1}{l|}{2.71e-10} & 1.12e-03                         \\
\textbf{Background Score}            & \multicolumn{1}{l|}{6.27e-07}  & 6.44e-03                         \\
\textbf{Scene Score}                 & \multicolumn{1}{l|}{3.15e-15} & 1.09e-05                       \\ \hline
\end{tabular}
\caption{\centering Results of a right-tailed paired t-test on the VLM-as-eval scores from 300 randomly sampled frames not in the train/test sets. Only compares statistical significance of the difference between the non-finetuned model and the finetuned model (no character detection is not in consideration). Even after applying conservative Bonferroni correction, we arrive at a p-value threshold of $\frac{0.05}{5} = 0.01$. Given this threshold, all differences between the quality of caption subsections with/without a finetuned character detection model are significant.}
\label{tab:llm_as_eval_pvalue}
\end{table}

\section{Discussion}
The experiments above provide corroborating pieces of evidence for two major claims. The first is that \textbf{SFT can be used to effectively align general-purpose VLMs for character identification with minimal overhead and allow smaller models to ``punch above their weight''}. The evaluated metrics of smaller models post-SFT is both better than off-the-shelf larger models and comparable to larger models post-SFT. [\ref{fig:f1_mistakes_plot}] shows this visually, and more comprehensive results are shown in [\ref{tab:main_metric_results}]. These results were also achieved using PeFT, not full-scale finetuning. In addition the training times for the 3B and 7B models were on the order of minutes with 8 $\times$ H100 Nvidia GPUs. This suggests that our work can be extended to resource-constrained settings as well and in industry settings is essentially no-cost. The second claim is that \textbf{using finetuned character detection as a step within a captioning pipeline improves holistic caption quality significantly}. This is described by [\ref{fig:llm_as_eval_charts}] succinctly. LLM-as-eval can excel in tasks such as text quality assessment, but it is also true that evaluation scores can be poorly aligned and that hallucinations affect the quality of these evaluations ~\citep{gu2025surveyllmasajudge}. To supplement our claim, we performed a paired t-test on the score differences between captions generated with an off-the-shelf character detection model and our finetuned model. With this evidence and the fact that the exact same model was used across all evaluations, we can claim that the differences indicate an improvement in holistic caption quality.

Is the character detection step necessary? [\ref{fig:llm_as_eval_charts}] shows that having no character detection is detrimental to caption quality. It could be said that the character detection step can be offloaded to the captioning model, thereby removing the need for a two-stage pipeline. This would entail giving quite a large context window to the captioning VLM. The issue arises in that excessively long context windows have been shown to harm LLM accuracy on reasoning tasks ~\citep{modarressi2025nolimalongcontextevaluationliteral}. By offloading a specific task to a separate model, we reduce the ``strain'' on the captioning LLM. In addition, more complex media may have dozens of characters. While character consistency is a very obvious axis to align a VLM on, one can extend this thinking to any important variable that can be annotated in a gold-standard dataset. It could be possible to explore the effectiveness of alignment on emotion, pose, camera angle, etc. We are unsure about the efficacy of creating an extended pipeline with models aligned on all these axes, and leave this to future work.

Underpinning all these improvements is the creation of a gold-standard dataset. This dataset defines any instance-grounded metrics, and therefore bias within the dataset may bias the calculated metrics to seem better or worse than their ``true'' value. Universal metrics like LLM-as-eval (model-based) or structural adherence (model-free) provide orthogonal pieces of evidence to see if image-caption pair alignment has truly improved. We suggest teams take care in curating a gold-standard dataset for alignment, with a simple yet effective prescription for this demonstrated in [\ref{fig:clip_embeds}]. We also hope teams can utilize the metric taxonomy provided in [\ref{fig:metric_taxonomy}] to think about what metrics are important, and when/where to use them.

We now take a look at a caption generated first with an off-the-shelf VLM as a baseline, and after with our two-stage captioning pipeline with a finetuned model Further examples can be found in [\ref{appendix:cap_improvements}].

\begin{figure}[H]
    \centering
    \texttt{Image:}\\
    \includegraphics[width=0.8\textwidth]{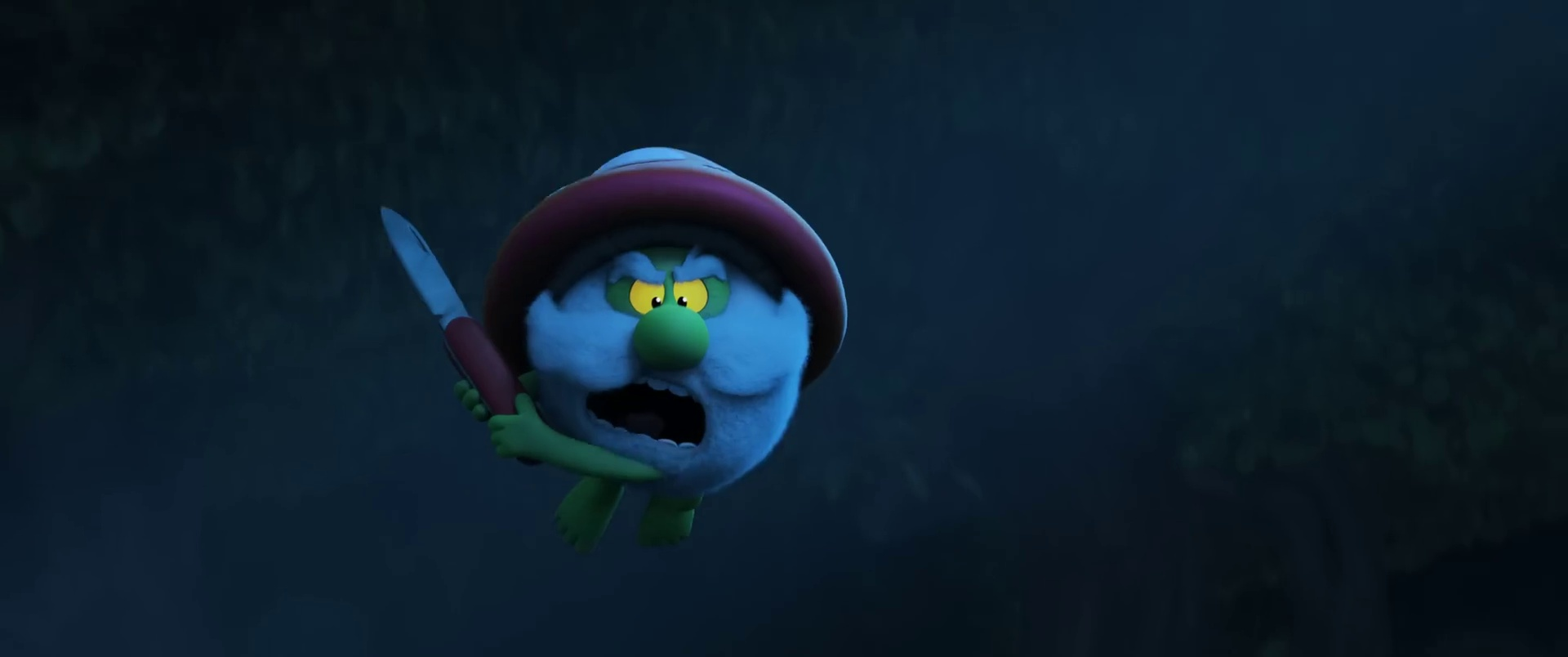}\\
    \texttt{Off-the-shelf captioning pipeline:}
    \begin{lstlisting}[style=caption]
        {
            "scene": "The image shows a small, green, mushroom-headed character named Sprite being held in a person's hand. Sprite is smiling and holding a knife with a red handle. The scene transitions to a darker setting where Sprite appears angry, with glowing yellow eyes and a fierce expression, still holding the knife.",
            "background": "The background is dark and blurred, suggesting a nighttime or dimly lit forest setting. The focus is on the character, with the background details being indistinct.",
            "characters": {
                "Sprite": {
                    "description": "Sprite is a small, green, humanoid character with a large mushroom cap for a head, wearing red shorts. The mushroom cap is red with white spots.",
                    "location": "In the first image, Sprite is in the center, being held in a person's hand. In the second image, Sprite is floating in the air.",
                    "expression": "In the first image, Sprite has a cheerful and happy expression. In the second image, Sprite has an angry and intense expression with glowing yellow eyes.",
                    "pose": "In the first image, Sprite is standing upright with one hand on the hip and the other holding a knife. In the second image, Sprite is floating with a knife in hand, mouth open as if shouting."
                }
            },
            "salient_objects": {
                "knife": "A knife with a red handle, held by Sprite in both images."
            }
        }
    \end{lstlisting}
    \texttt{Improved captioning pipeline:}
    \begin{lstlisting}[style=caption]
        {
            "scene": "Elder_Sprite is depicted in a dark, ominous setting, holding a knife with a determined and fierce expression.",
            "background": "The background is dark and shadowy, creating a mysterious and tense atmosphere.",
            "characters": {
                "Elder_Sprite": {
                    "description": "Elder_Sprite is a round, green-faced character with a white beard and a red mushroom cap. The character has a green nose and green hands.",
                    "location": "Center of the image",
                    "expression": "Angry and intense, with glowing yellow eyes and an open mouth as if shouting",
                    "pose": "Holding a knife in a defensive or aggressive stance"
                }
            },
            "salient_objects": {
                "knife": "A knife with a red handle, held by Elder_Sprite"
            }
        }
    \end{lstlisting}
    \label{fig:compare_caption_3}
\end{figure}

Not only was the character misidentified in the baseline caption, but the model seems to confuse its given context with the image it was asked to annotate. In contrast, the description generated with the improved pipeline is succinct and contains far fewer hallucinations. It has been shown that throughout a whole training dataset, a constant number of poisoned examples can lead to degraded model performance ~\citep{souly2025poisoningattacksllmsrequire}. Although poorly aligned image-caption pairs do not qualify as malicious poisoning, the idea remains the same: data quality matters more than one may think. It is therefore paramount to focus on image-caption alignment when training a T2I model. The findings demonstrated in this paper are one way to improve this alignment.

What about prompt engineering? As was discovered through this project, the quality of VLM outputs is tied intimately to prompt structure. We used model-specific provided prescriptions and common knowledge about prompt engineering for each prompt we crafted. This was to improve output quality and ensure we are giving each VLM the ability to perform at near-maximum potential. All used prompts can be found in [\ref{appendix:vlm_prompts}].

\section{Conclusion}

Data quality control is becoming an evermore important aspect of the end-to-end large model training paradigm. A large subset of these models, text-to-image models, require image-caption pairs where the caption accurately describes the important features of the image. We have demonstrated here that an off-the-shelf captioning model comes with significant issues, namely that characters are often misidentified which can have unforeseen downstream consequences for caption quality. To remedy this, we prescribe a simple two-stage captioning improvement pipeline that has been shown to statistically improve caption quality across the board. Furthermore, by also using our metric taxonomy, teams can leverage simple SFT to greatly improve the quality of their image-caption pairs on an axis of their choosing. We re-iterate the importance of creating a diverse gold-standard dataset as an anchor for important metrics and alignment, and suggest teams use LabelStudio and VLM pre-annotation to accelerate the process and save precious developer hours.

\section*{Acknowledgments}
Thank you to my peers at Adobe that made this paper possible by assisting with ideation, data collection, wrangling, training, and paper writing.

\bibliographystyle{plainnat}
\bibliography{references}

\appendix

\section{SFT Training Curves} \label{appendix:train_curves}
\begin{figure}[H]
    \centering
    \begin{table}[H]
    \centering
    \begin{tabular}{cc}
        \includegraphics[width=0.45\textwidth]{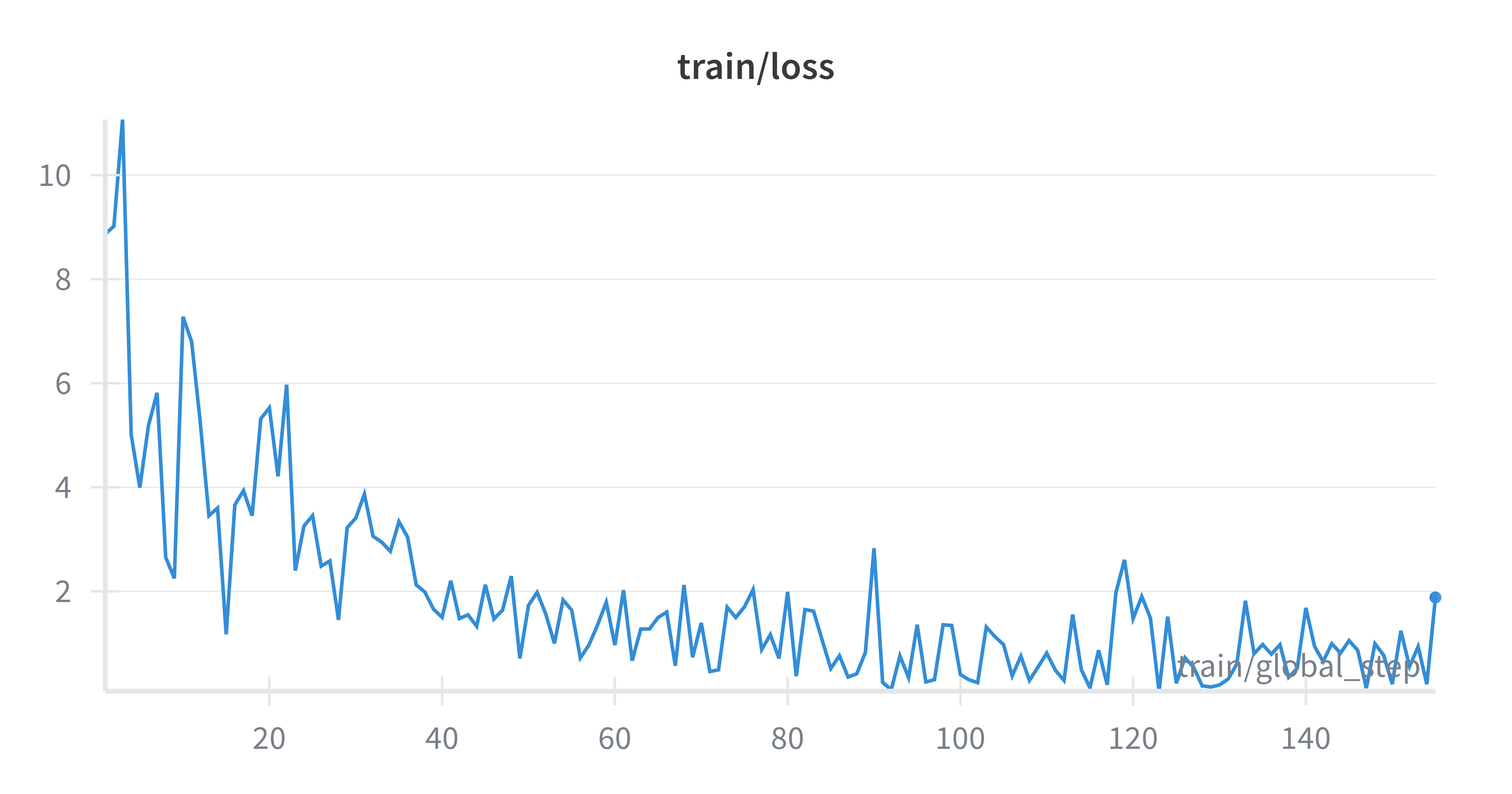} &
        \includegraphics[width=0.45\textwidth]{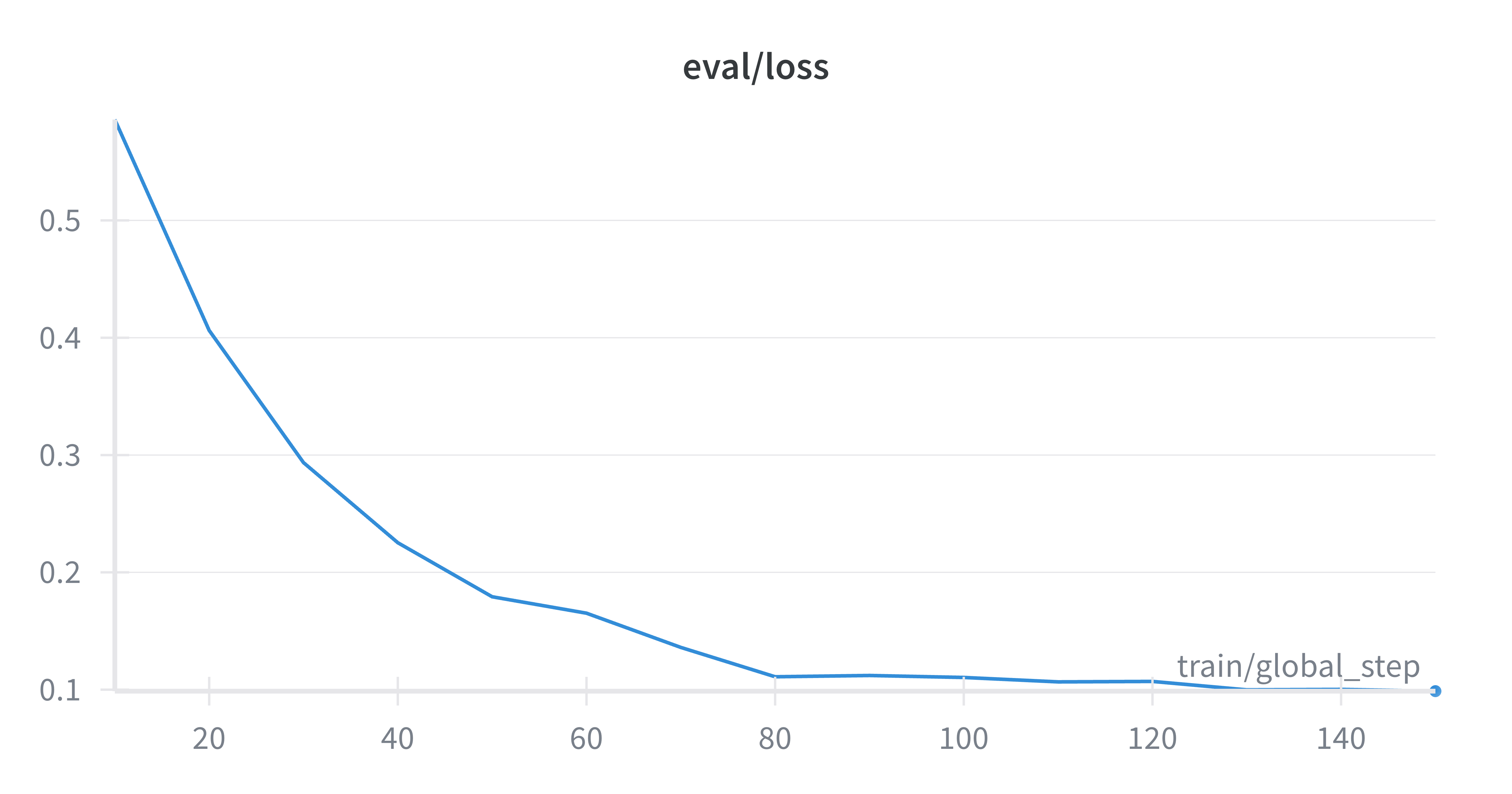} \\
        \includegraphics[width=0.45\textwidth]{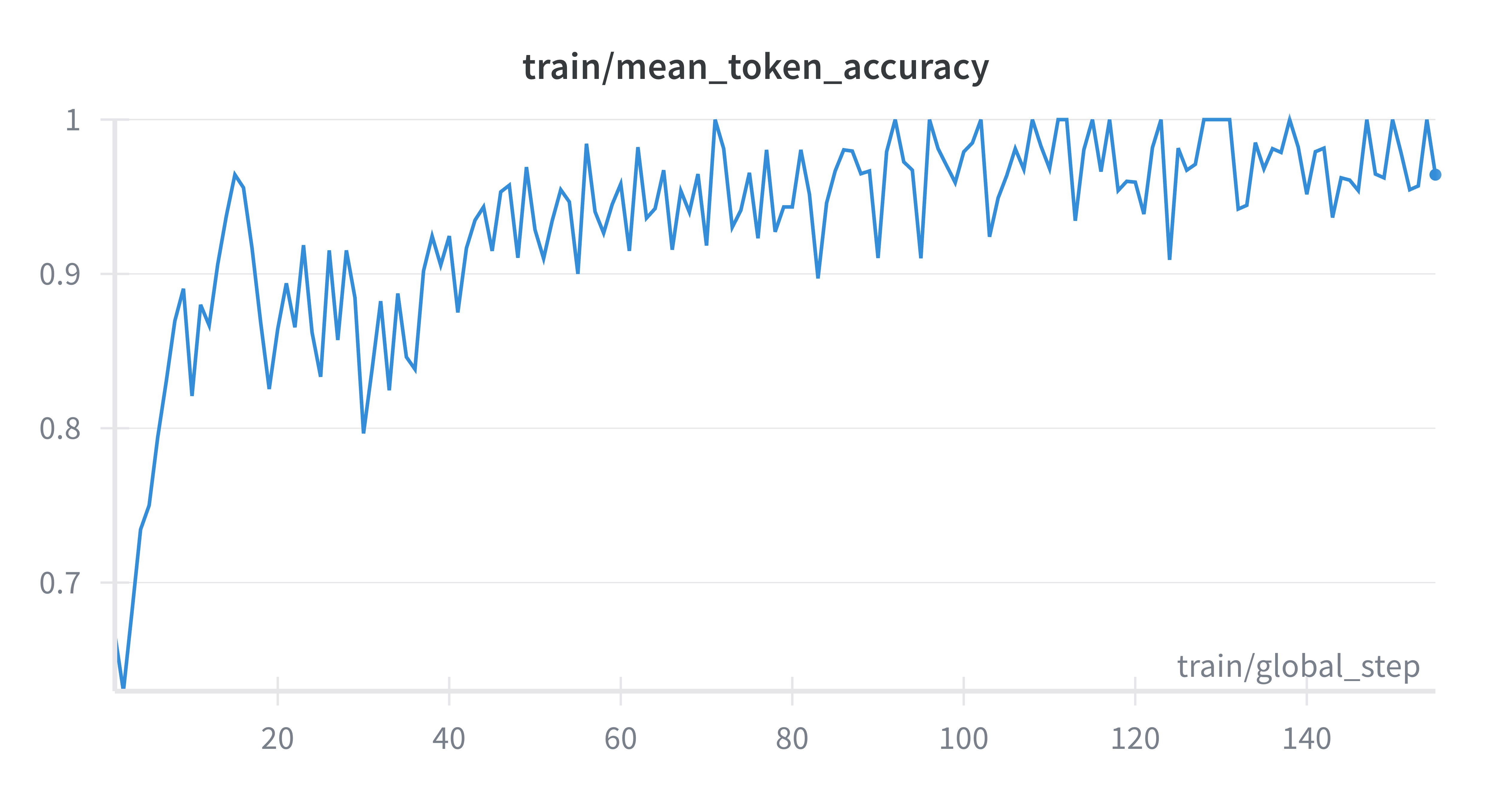} &
        \includegraphics[width=0.45\textwidth]{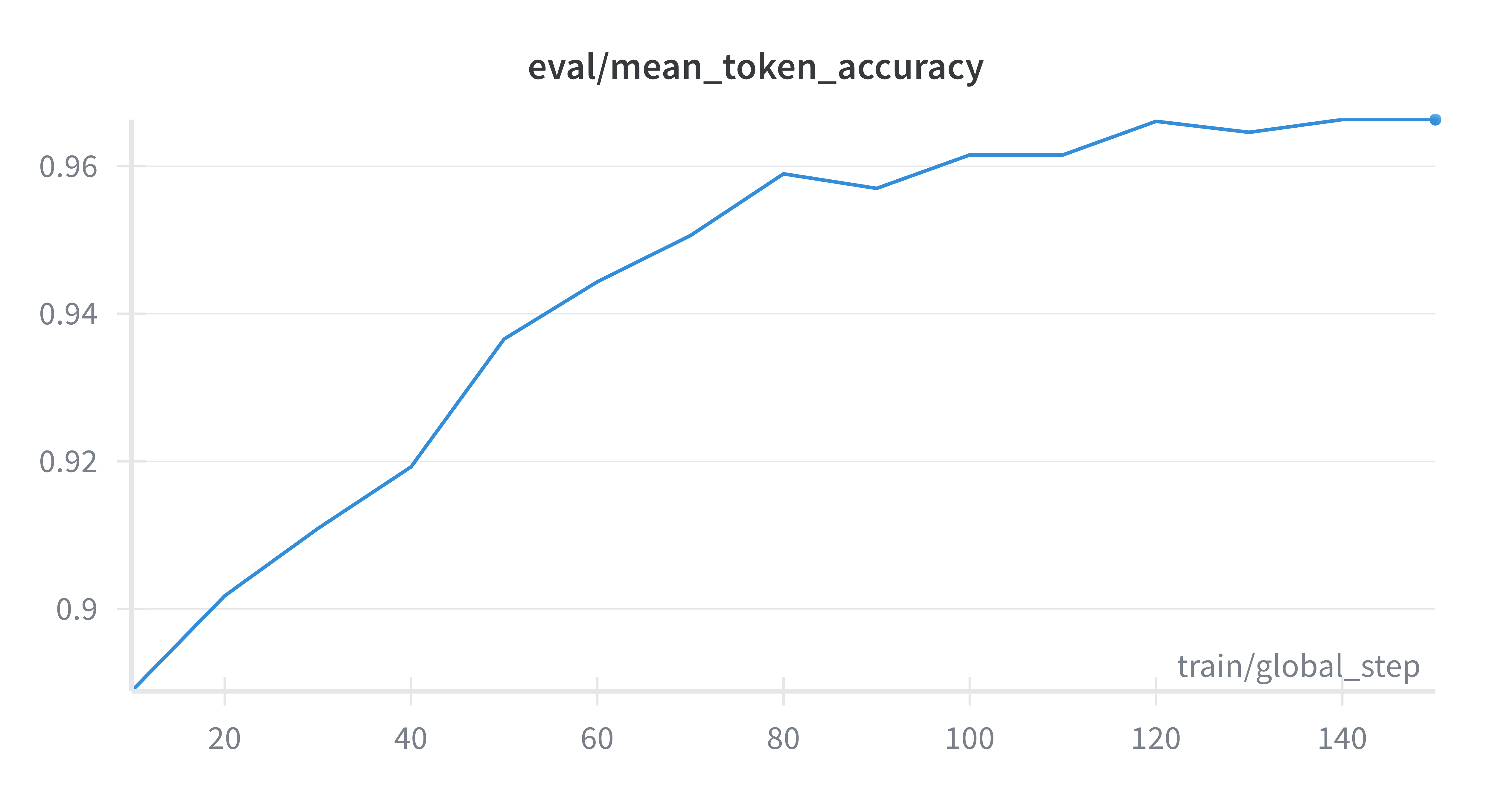} \\
    \end{tabular}
    \end{table}
\caption{\centering Training/Evaluation loss and mean token accuracy curves for finetuning Qwen2.5-VL-3B. Model trained using PEFT (LoRA) for 5 epochs and a learning rate of 1e-4 with the AdamWFused optimizer.}
\label{fig:3b_wandb}
\end{figure}

\begin{figure}[H]
    \centering
    \begin{table}[H]
    \centering
    \begin{tabular}{cc}
        \includegraphics[width=0.45\textwidth]{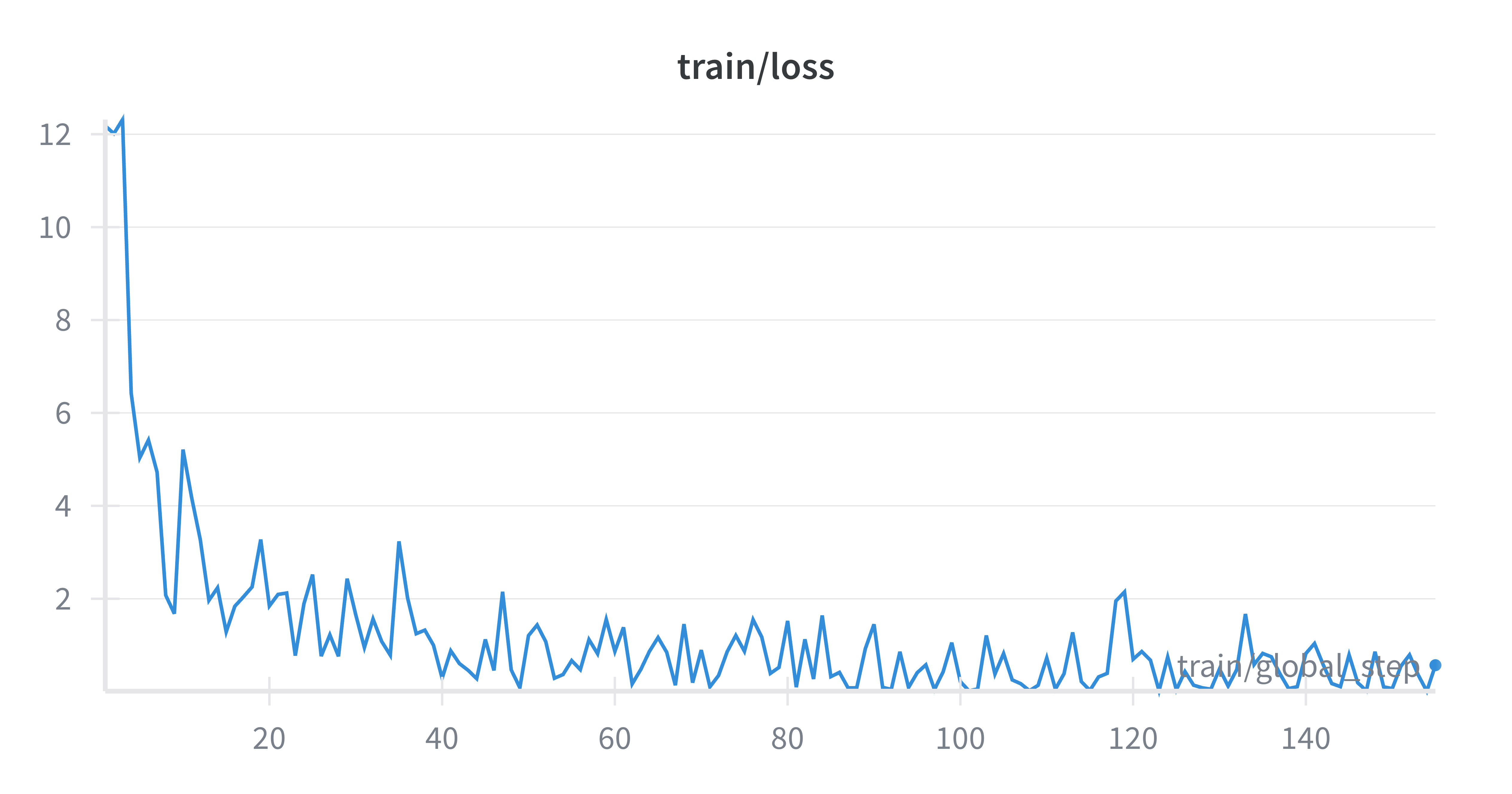} &
        \includegraphics[width=0.45\textwidth]{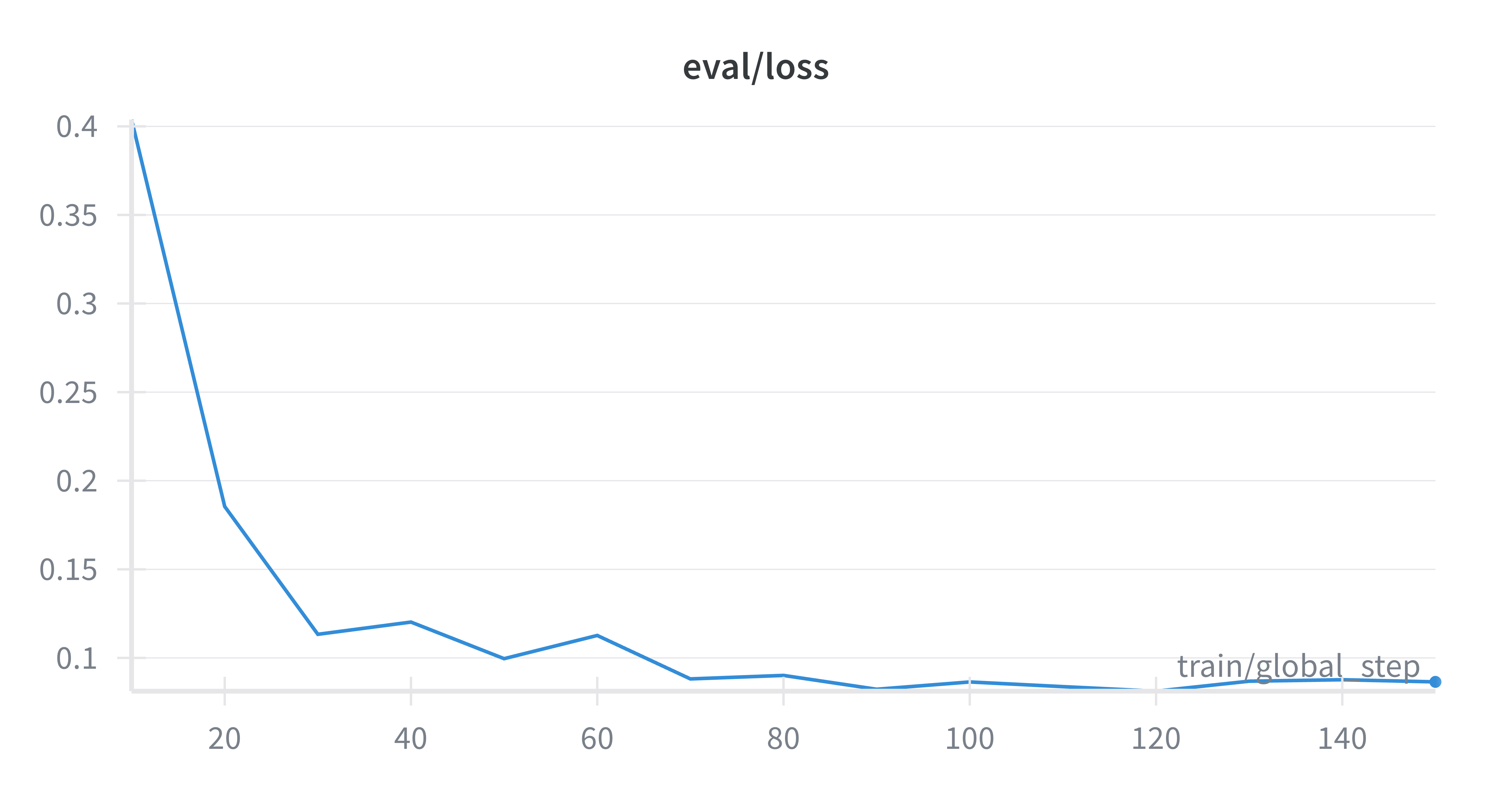} \\
        \includegraphics[width=0.45\textwidth]{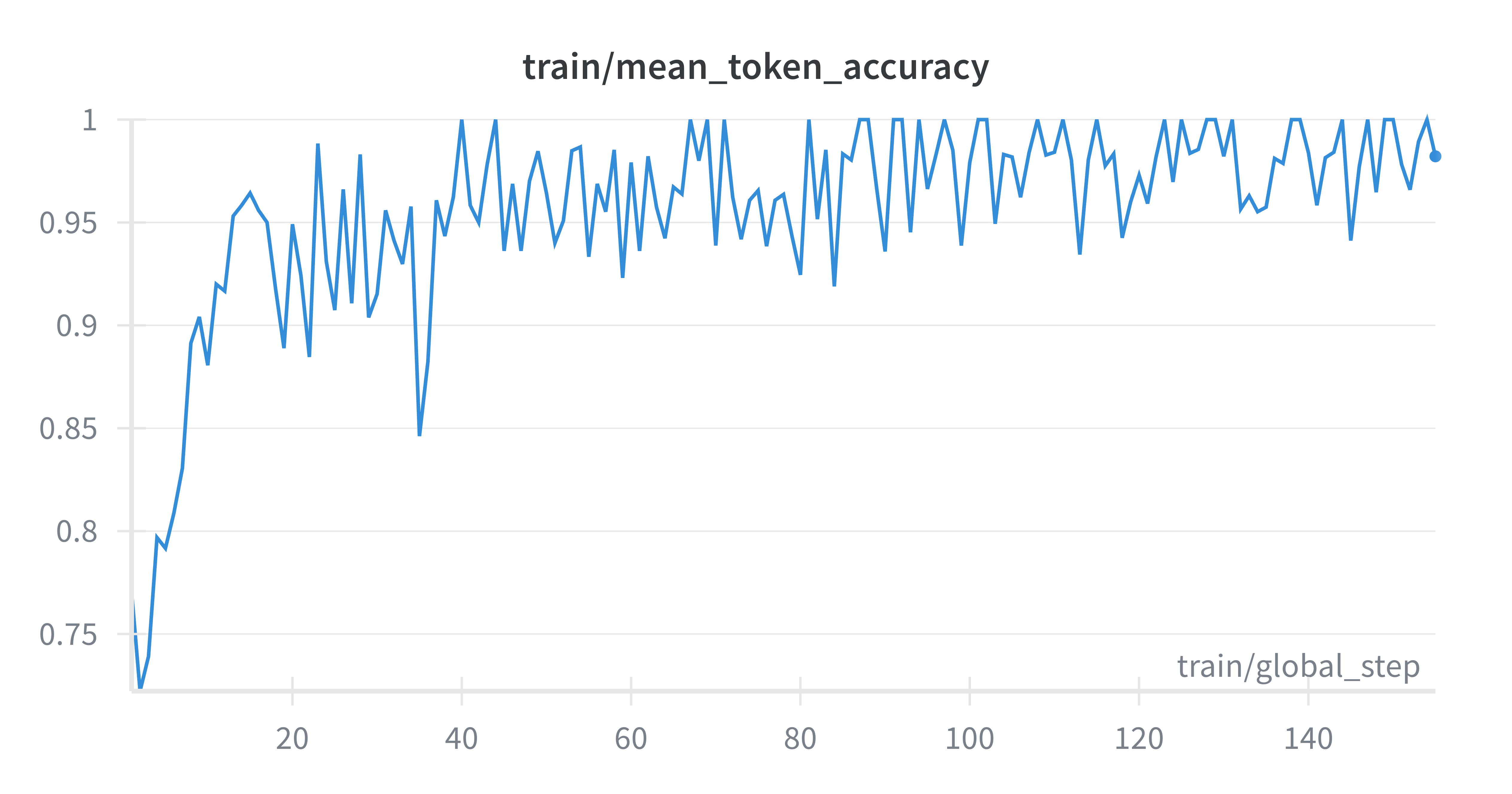} &
        \includegraphics[width=0.45\textwidth]{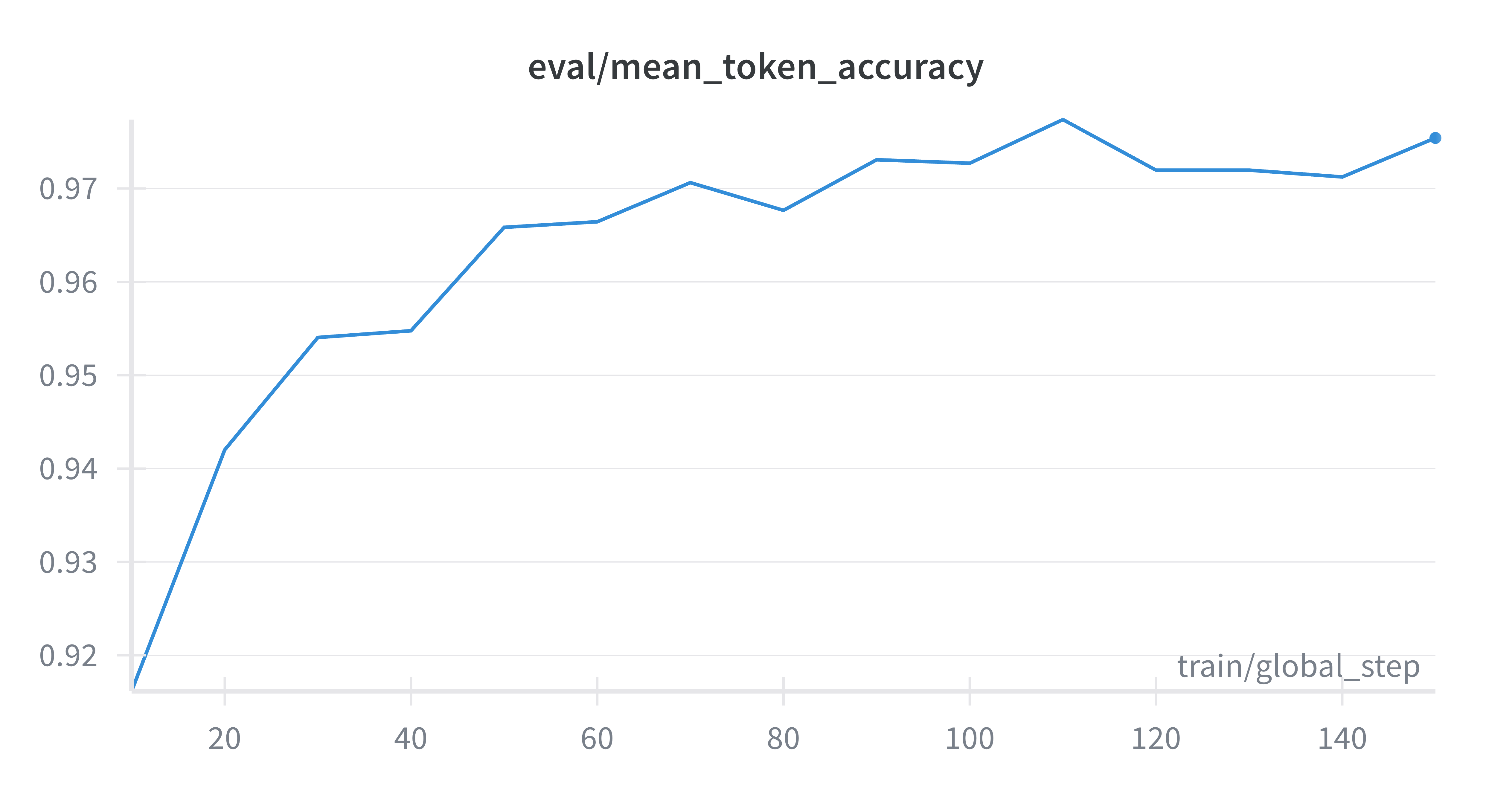} \\
    \end{tabular}
    \end{table}
\caption{\centering Training/Evaluation loss and mean token accuracy curves for finetuning Qwen2.5-VL-7B. Model trained using PEFT (LoRA) for 5 epochs and a learning rate of 1e-4 with the AdamWFused optimizer.}
\label{fig:7b_wandb}
\end{figure}

\begin{figure}[H]
    \centering
    \begin{table}[H]
    \centering
    \begin{tabular}{cc}
        \includegraphics[width=0.45\textwidth]{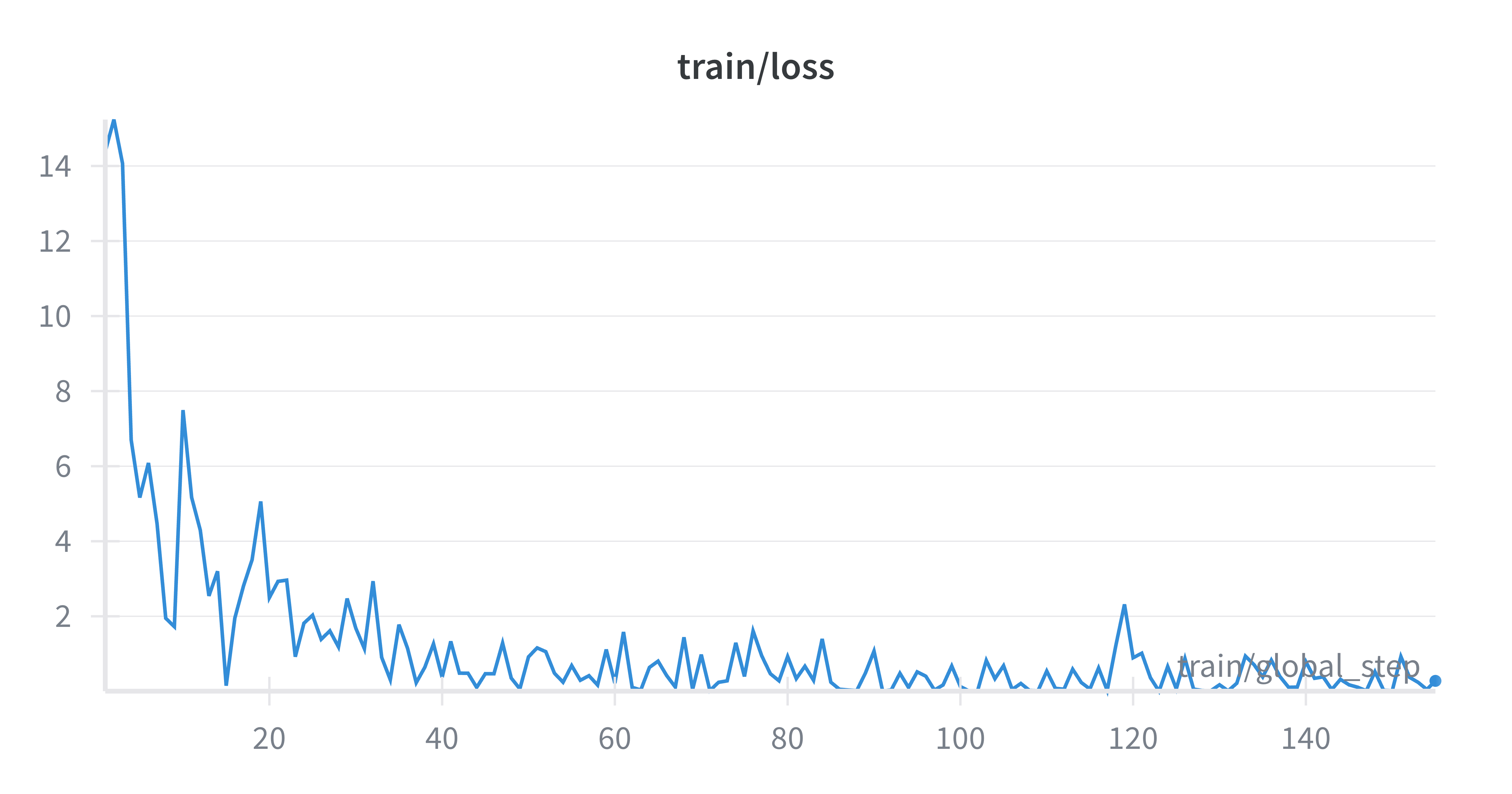} &
        \includegraphics[width=0.45\textwidth]{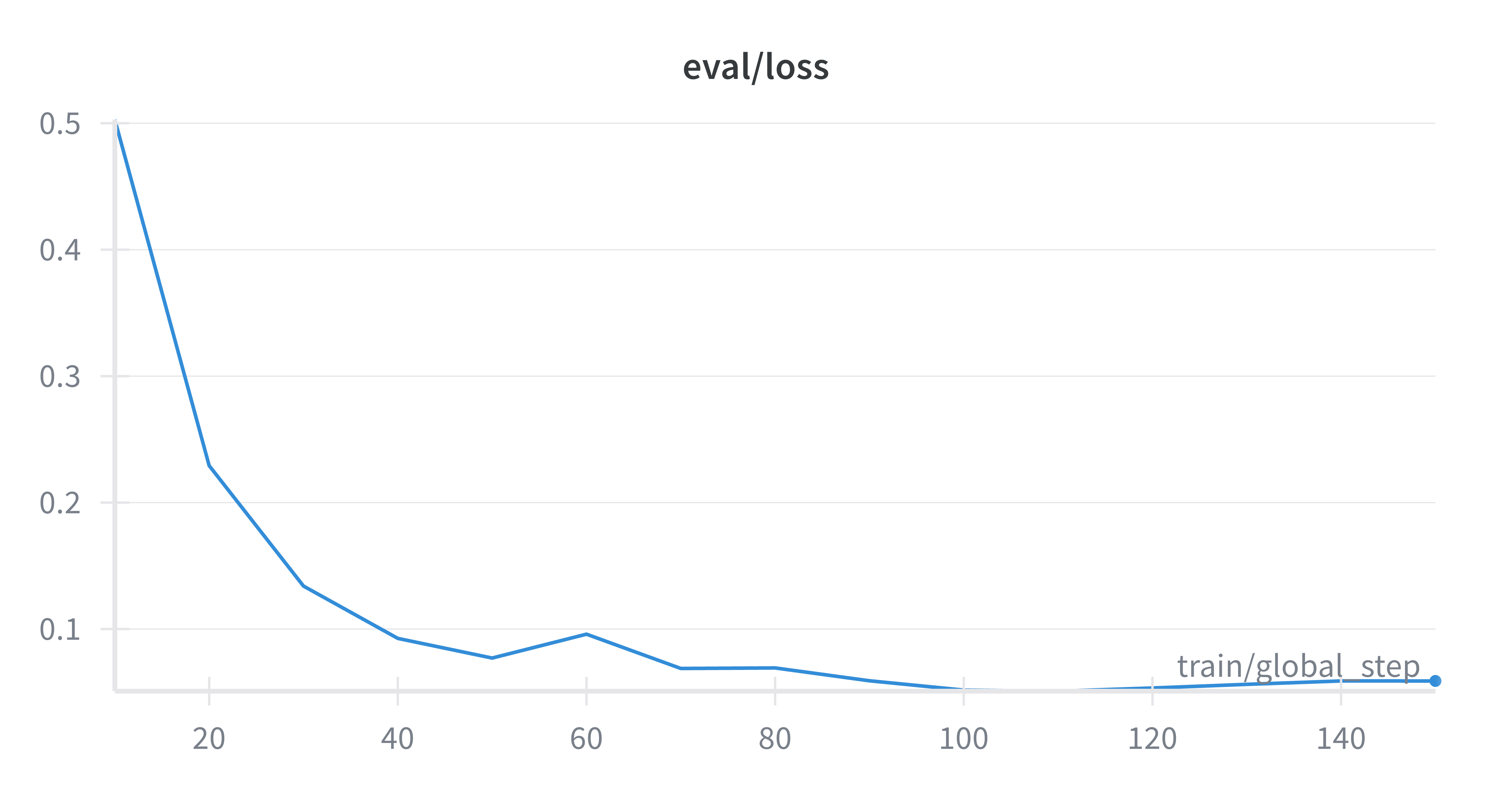} \\
        \includegraphics[width=0.45\textwidth]{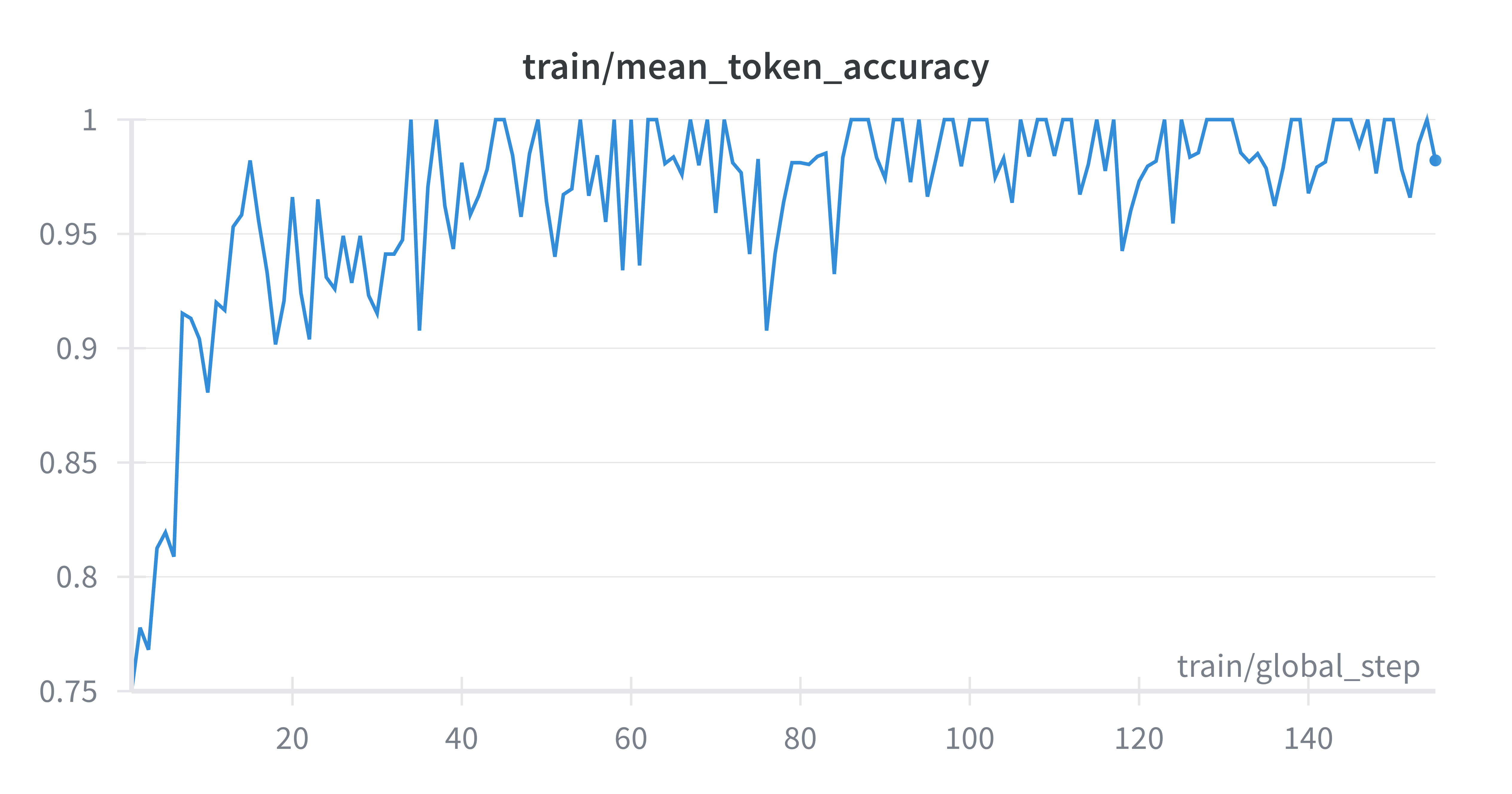} &
        \includegraphics[width=0.45\textwidth]{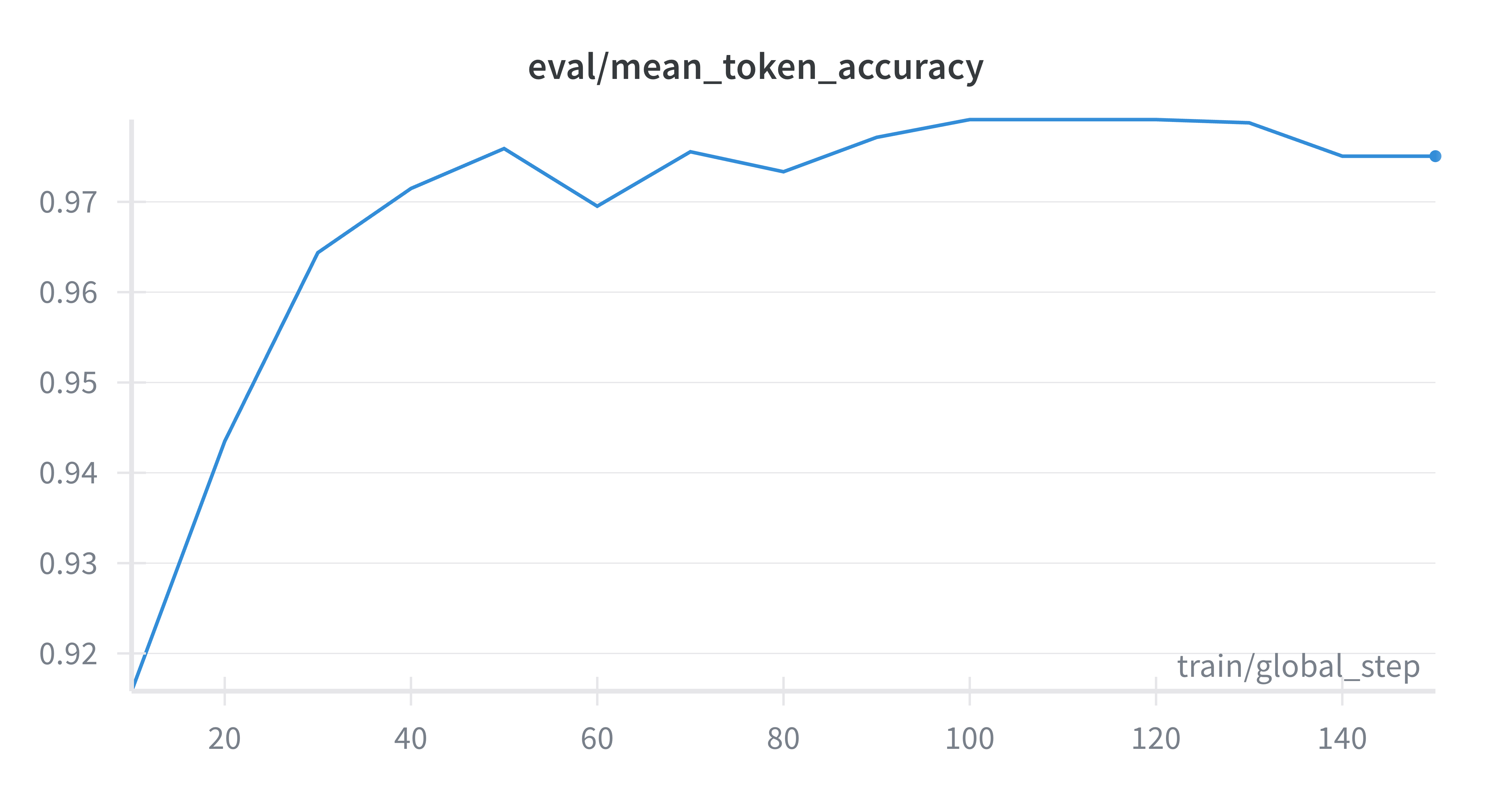} \\
    \end{tabular}
    \end{table}
\caption{\centering Training/Evaluation loss and mean token accuracy curves for finetuning Qwen2.5-VL-32B. Model trained using PEFT (LoRA) for 5 epochs and a learning rate of 1e-4 with the AdamWFused optimizer.}
\label{fig:32b_wandb}
\end{figure}

\section{VLM Prompts} \label{appendix:vlm_prompts}
The following VLM prompts all have interleaved text and images, allowing for the VLM to have a richer understanding of the universe in which its tasks lie. An important part of all prompts is the character context section, where images of each character (see [\ref{fig:character_grid}]) are provided along with their name and possibly a short description. If this section has all, some, or none of the characters depends on the context of the VLM call.

\begin{lstlisting}[style=llmprompt,caption={\centering Qwen2.5-VL prompt for character detection, both for SFT and as part of the captioning pipeline. Every possible character and their names are provided here as context. The final image is the one the VLM is tasked with detecting characters in.}]
SYSTEM PROMPT = You are an expert in character detection. You are given a list of POTENTIAL characters present in the image. You will then be given a target image, and you will need to provide a list of characters that are definitely present in the image. Use these images to help you understand what each character looks like.

USER PROMPT = Given the solo images of the characters that COULD be present in the image, please provide a list of characters that are definitely present in the target image. A list of valid characters is provided below. 
["Ellie", "Jay", "Phil", "Rex", "Victoria", "Sprite", "Elder_Sprite"].
Think through your output carefully and do not make assumptions. Your output should be a list of characters in the format: ["character1", "character2", "character3"] with no other text.Think step by step and do not make assumptions. Do not hallucinate characters that are not in the list.

prompt = [
    {"role": "system", "content": [{"type": "text", "text": SYSTEM_PROMPT}]},
    {
        "role": "user",
        "content": [
            {"type": "text", "text": "===POTENTIAL CHARACTERS HERE==="},
            *(
                item
                for _, text in character_sheet
                for item in (
                    {"type": "image"},
                    {"type": "text", "text": text},
                )
            ),
            {"type": "text", "text": "===POTENTIAL CHARACTERS HERE==="},
        ],
    },
    {
        "role": "user",
        "content": [
            {"type": "text", "text": "===TARGET IMAGE HERE==="},
            {"type": "image"},
            {"type": "text", "text": "===TARGET IMAGE HERE==="},
        ],
    },
    {"role": "user", "content": [{"type": "text", "text": USER_PROMPT}]},
]
\end{lstlisting}

\begin{lstlisting}[style=llmprompt,caption={\centering InternVL-38B prompt to caption images from the dataset. Caption output is structured. Information about what subset of characters are in the image is provided by the Qwen2.5-VL character detection model. The final image is the one the VLM is tasked with captioning.}]
You are a helpful assistant that can describe images in a structured way. You are given an image and characters present in the image. Please provide a description for the given image as if the description is used to prompt a text to image generator model to recreate the image.

== CHARACTER CONTEXT ===
<image1>
<desc1>
<image2>
<desc2>
...
=== END CHARACTER CONTEXT ===
=== INSTRUCTIONS ===
The description should cover the following aspects and categories:

Scene Description: Describe the scene in detail, use the character names from the character images provided if seen in the image. 
If no characters are present then just describe the scene and background.
Describe how the characters are interacting with each other.

Background: Describe the setting and background, mentioning everything in the image.

Characters: Details of characters present in the image. 
Give a structured description of each character with these fields for each character:
    1. Description: A detailed description of the character.
    2. Location: The location of the character in the image.
    3. Expression: The expression of the character.
    4. Pose: The pose of the character.
If none of the characters are present, make this field an empty string.

Salient Objects: Provide a list of salient objects present in the image. A dictionary with object name as key and object description as value.

The output should strictly follow the JSON format. Do not use markdown code blocks, just output the raw JSON object.
{{
    "scene": "Scene Description with characters and their respective actions and poses",
    "background": "Details of background describing everything in the image",
    "characters": "Details of characters present in the image, this is a dictionary with each key being the character name and the value being a structured dictionary with the character's description, expression, pose, and if not sure about the character, use the name of the character as 'Unknown Character'",
    "salient_objects": "List of salient objects present in the image, a dictionary with object name as key and object description as value"
}}
=== END INSTRUCTIONS ===
=== IMAGE TO CAPTION ===
<image>
\end{lstlisting}

\begin{lstlisting}[style=llmprompt,caption={\centering Gemini-2.5-Pro prompt for holistic evaluation of each subsection of the structured caption. The overall score displayed in above plots comes from a simple mean of all section scores. Every possible character and their names are provided here as context. The alignment between the final image and structured caption (generated by some Qwen2.5-VL character detector + InternVL3-38B) is what this VLM is determining the quality of.}]
You are a senior VLM quality evaluator. You will be shown reference character images and names from this show to aid identity verification only. Use these references solely to check whether the candidate caption names characters correctly and to avoid hallucinated identities. Do not copy descriptions from the references or introduce details not visible in the evaluated image. When uncertain about identity, prefer 'Unknown Character'.
=== CHARACTER CONTEXT ===
<image1>
<desc1>
<image2>
<desc2>
...
=== END CHARACTER CONTEXT ===
You are evaluating how well a candidate caption aligns with the image content. Your task is not to rewrite the caption, but to grade its evidence-grounded quality holistically.

Strict grounding and adjudication policy:
- Ground every judgment FIRST on the main image, THEN on the candidate caption; use character references only for identity verification (no new details).
- If a claim in the caption is not visible or is contradicted by the image, deduct points (hallucination).
- Do NOT penalize schema/formatting/style inconsistencies of the caption (keys, headings, ordering). Judge the information content only. If a nominal "field" is missing, infer from available content where possible and score based on substance.
- Do not use external/world knowledge; judge only what the image shows and the caption asserts.
- Prefer precision over verbosity; penalize vague language.
- Be extra strict about character correctness: missing visible characters or mislabeled identities anywhere in the caption (scene, background, characters, salient_objects) should incur heavy deductions.

Aspects to evaluate (aligned with the captioning prompt):
- Scene Description (key: "scene")
  1) Identify which characters are present and where they are located. Use character names from the character images provided if they are clearly visible in the image.
  2) Describe what each character is doing (actions and poses) and their interactions.
  3) Capture spatial relationships and notable expressions.
  4) Explicitly acknowledge all visible characters; omission of a clearly visible character is a severe error.
  5) If no characters are present, describe the scene and background only.
  6) If unsure about character identity, use 'Unknown Character'. Do NOT add characters not present.

- Background (key: "background")
  - Describe the setting and background, grounded in what is visible (environment type, notable structures/props/signage, lighting/time-of-day/visual cues) with relevant specifics.

- Characters (key: "characters")
  - For each visible character: include a short description, location (left/right/center/foreground/background, relative positions), facial expression, and pose/action.
  - Identity handling: verify with provided references when possible; prefer 'Unknown Character' over guessing and justify identity with visible attributes.
  - Capture interactions (gaze, touch, shared activity) and spatial relations when evident.
  - Treat omissions of visible characters or incorrect/mislabeled identities as severe errors.

- Salient Objects (key: "salient_objects")
  - Include key non-character objects that meaningfully contribute to understanding; briefly note attributes that matter (color, state, quantity, relation to characters/background). Avoid hallucinations and trivial lists.

Scoring rubric (integers 1-10 per section, balanced and evidence-based):
Evaluate each section holistically across three dimensions, then blend them:
- Evidence-grounded correctness (50%): Are claims supported by the image and free of hallucinations?
- Completeness and specificity (30%): Are salient details and relationships covered with sufficient precision?
- Clarity and concision (20%): Is the information communicated clearly and succinctly? Ignore caption schema/formatting when scoring; focus on content quality.

Use anchors to guide scoring:
- 9-10 (Excellent): Strong on all three dimensions; precise and comprehensive; no material errors.
- 7-8 (Strong): Well-grounded with minor omissions or small imprecision; clearly image-based.
- 5-6 (Adequate): Generally grounded with a few moderate issues or multiple minor ones.
- 3-4 (Limited): Noticeable gaps in grounding or completeness; repeated vagueness.
- 1-2 (Poor): Major grounding errors, invented details, or missing required structure/content.

Calibration guidance:
- Typical good-but-imperfect captions fall around 5-6; reserve 8-9 for near-exemplary work; 10 is rare.
- Award partial credit within each dimension rather than all-or-nothing penalties.
- Down-weight only the relevant dimension when a constraint is slightly missed.

Deduction and partial credit (apply consistently):
- Start each section at 10 while considering the three dimensions; subtract proportionally:
  - Minor issue: -0.5 to -1 (small imprecision, slight omission, mild verbosity).
  - Moderate issue: -1 to -2 (clear but contained omissions or ambiguous descriptions).
  - Hard violation: -3 (hallucinated identity/object; confident claims contradicted by the image; pervasive vagueness).
  - Character omissions/mislabels: -2 to -3 per affected character (-3 for missing a clearly visible or primary character; -3 for mislabeled identity contradicted by evidence).
- Do not double-penalize the same issue; if pervasive, apply once per distinct aspect (e.g., per character or object type) and reflect scope in the deduction magnitude.
- Clamp each section score to the [1, 10] range.

Section-specific notes for scoring (apply the 50/30/20 weighting):
- scene_score: Prioritize correctness of who/where/what; include spatial relations, actions, interactions, and expressions; ensure all visible characters are acknowledged; use provided names when justified; otherwise 'Unknown Character' is acceptable. Missing or mislabeled characters should be penalized strongly.
- background_score: Reward accurate, relevant setting details (lighting, structures, environment) grounded in the image; no penalty for omitting trivial elements.
- characters_score: Reward correct identities or justified uncertainty, specific locations, expressions, poses, and interactions. Strongly penalize missing visible characters and mislabeled identities.
- salient_objects_score: Reward inclusion of the most meaningful non-character objects with brief, correct attributes; penalize hallucinations more than omissions.

Output requirements:
- Output a single JSON object only. No markdown. Keys and types:
{
    "scene_score": <int 1-10>,
    "background_score": <int 1-10>,
    "characters_score": <int 1-10>,
    "salient_objects_score": <int 1-10>,
    "rationale": "Brief justification (2-4 sentences) explicitly citing what is or isn't visible in the image"
}
"""=== IMAGE TO EVALUATE ==="""
<image>
"""=== CANDIDATE CAPTION TO EVALUATE ==="""
<structured_caption>
\end{lstlisting}

\section{Example Caption Improvements} \label{appendix:cap_improvements}

\begin{figure}[H]
    \centering
    \texttt{Image:}\\
    \includegraphics[width=0.8\textwidth]{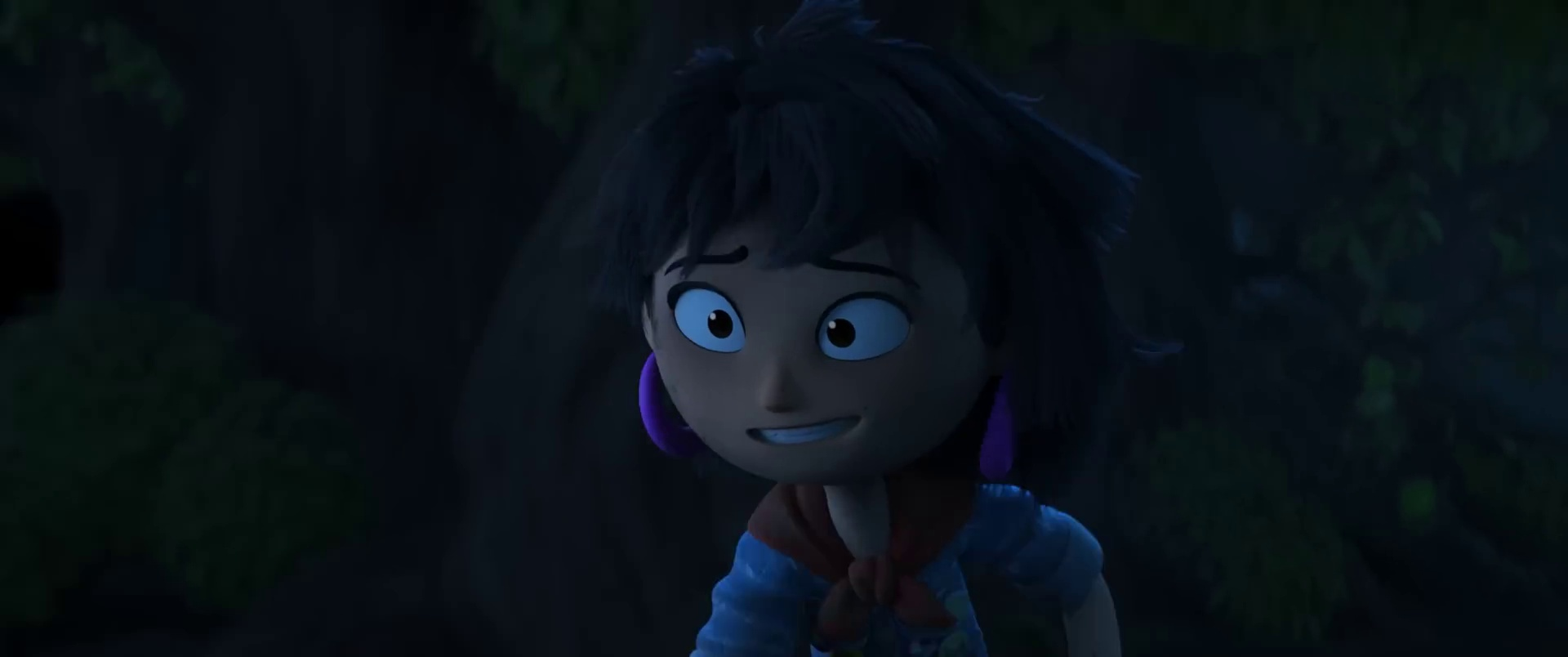}\\
    \texttt{Off-the-shelf captioning pipeline:}
    \begin{lstlisting}[style=caption]
        {
            "scene": "The image shows a character named Victoria in a dark, forest-like setting. She appears to be looking forward with a concerned expression.",
            "background": "The background is dark and shadowy, with hints of foliage, suggesting a nighttime forest or cave setting.",
            "characters": {
                "Victoria": {
                    "description": "Victoria has dark hair, large expressive eyes, and is wearing a blue shirt with a red scarf and purple earrings.",
                    "location": "Center of the image",
                    "expression": "Concerned",
                    "pose": "Standing and looking forward"
                }
            },
            "salient_objects": {}
        }
    \end{lstlisting}
    \texttt{Improved captioning pipeline:}
    \begin{lstlisting}[style=caption]
        {
            "scene": "Ellie is standing in a dark, forested area, looking slightly concerned or curious. She is the main focus of the image, with her expression and pose suggesting she is cautiously observing something.",
            "background": "The background is dark and shadowy, with hints of foliage and trees, indicating a nighttime forest setting.",
            "characters": {
                "Ellie": {
                    "description": "Ellie has dark hair tied in a ponytail with a blue scrunchie, large purple hoop earrings, and is wearing a denim jacket with a red scarf, khaki shorts, and red boots. She has a colorful fanny pack around her waist.",
                    "location": "Center of the image",
                    "expression": "Concerned or curious",
                    "pose": "Standing with a slight forward lean, looking intently ahead"
                }
            },
            "salient_objects": {}
        }
    \end{lstlisting}
    \label{fig:compare_caption_1}
\end{figure}

\begin{figure}[H]
    \centering
    \texttt{Image:}\\
    \includegraphics[width=0.8\textwidth]{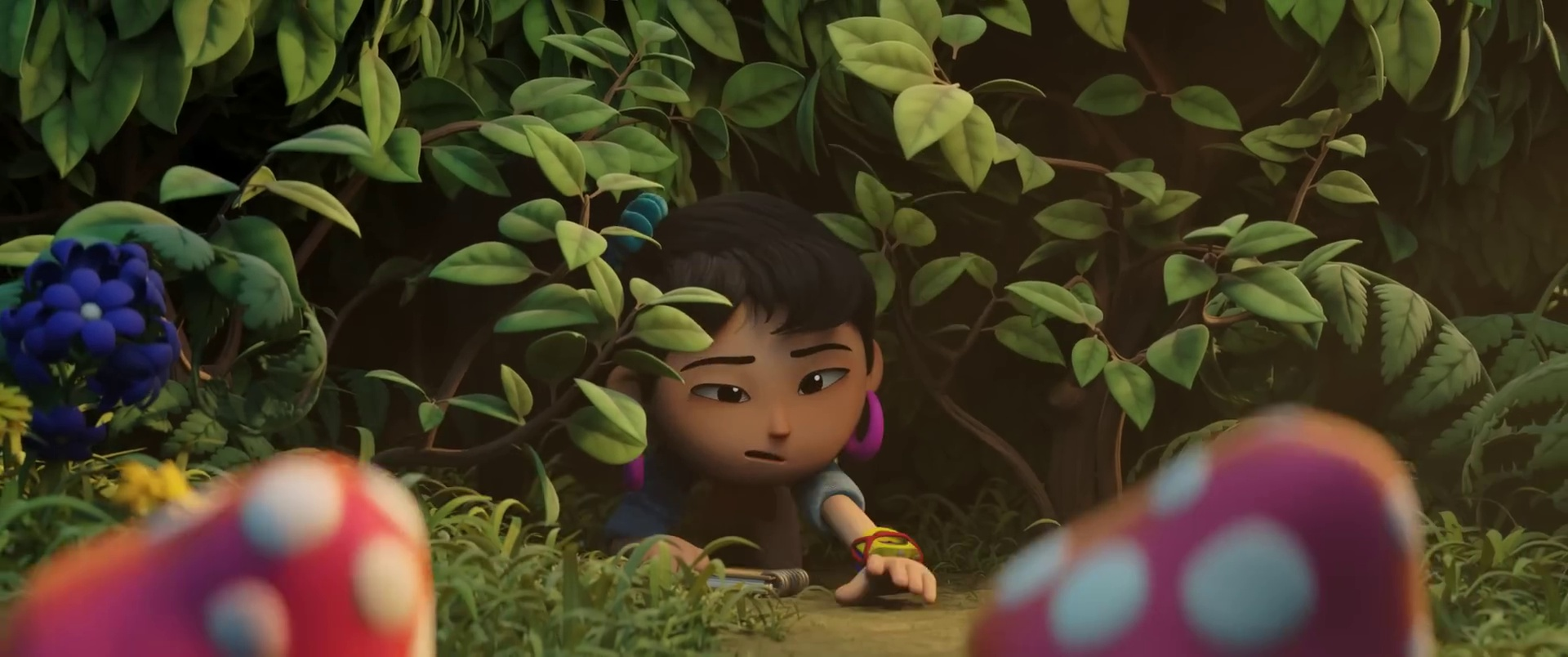}\\
    \texttt{Off-the-shelf captioning pipeline:}
    \begin{lstlisting}[style=caption]
        {
            "scene": "Victoria is crouching in a lush, green forest, looking intently at something in front of her. She appears cautious and curious, with a focused expression.",
            "background": "The scene is set in a dense, vibrant forest with various green leaves and plants. There are colorful flowers and large mushrooms in the foreground.",
            "characters": {
                "Victoria": {
                    "description": "Victoria has short dark hair with a blue hair accessory, wears a blue top, and has pink earrings. She is adorned with colorful bracelets.",
                    "location": "Crouching among the foliage in the center of the image.",
                    "expression": "Focused and slightly concerned.",
                    "pose": "Crouching on the ground, leaning forward with one hand on the ground and the other holding a small object."
                }
            },
            "salient_objects": {
                "Mushrooms": "Large mushrooms with red caps and white spots in the foreground.",
                "Blue Flowers": "Cluster of blue flowers to the left of Victoria.",
                "Green Foliage": "Dense green leaves and plants surrounding Victoria."
            }
        }
    \end{lstlisting}
    \texttt{Improved captioning pipeline:}
    \begin{lstlisting}[style=caption]
        {
            "scene": "Ellie is crouching in a lush, green forest, looking intently at something in front of her. She appears cautious and curious, with a focused expression on her face.",
            "background": "The background is a dense forest with various green plants and leaves. There are colorful flowers and large mushrooms visible, adding to the vibrant and whimsical atmosphere.",
            "characters": {
                "Ellie": {
                    "description": "Ellie has dark hair tied in a ponytail with a blue scrunchie, wearing a denim jacket with a red scarf, khaki shorts, and red boots. She has large pink hoop earrings and a colorful wristband.",
                    "location": "Crouching on the ground amidst the foliage",
                    "expression": "Focused and slightly concerned",
                    "pose": "Crouching with one hand on the ground and the other holding a notebook"
                }
            },
            "salient_objects": {
                "Blue Flowers": "A cluster of vibrant blue flowers to the left of Ellie",
                "Pink Mushrooms": "Large pink mushrooms with white spots in the foreground",
                "Notebook": "A small notebook held by Ellie"
            }
        }
    \end{lstlisting}
    \label{fig:compare_caption_2}
\end{figure}

\begin{figure}[H]
    \centering
    \texttt{Image:}\\
    \includegraphics[width=0.8\textwidth]{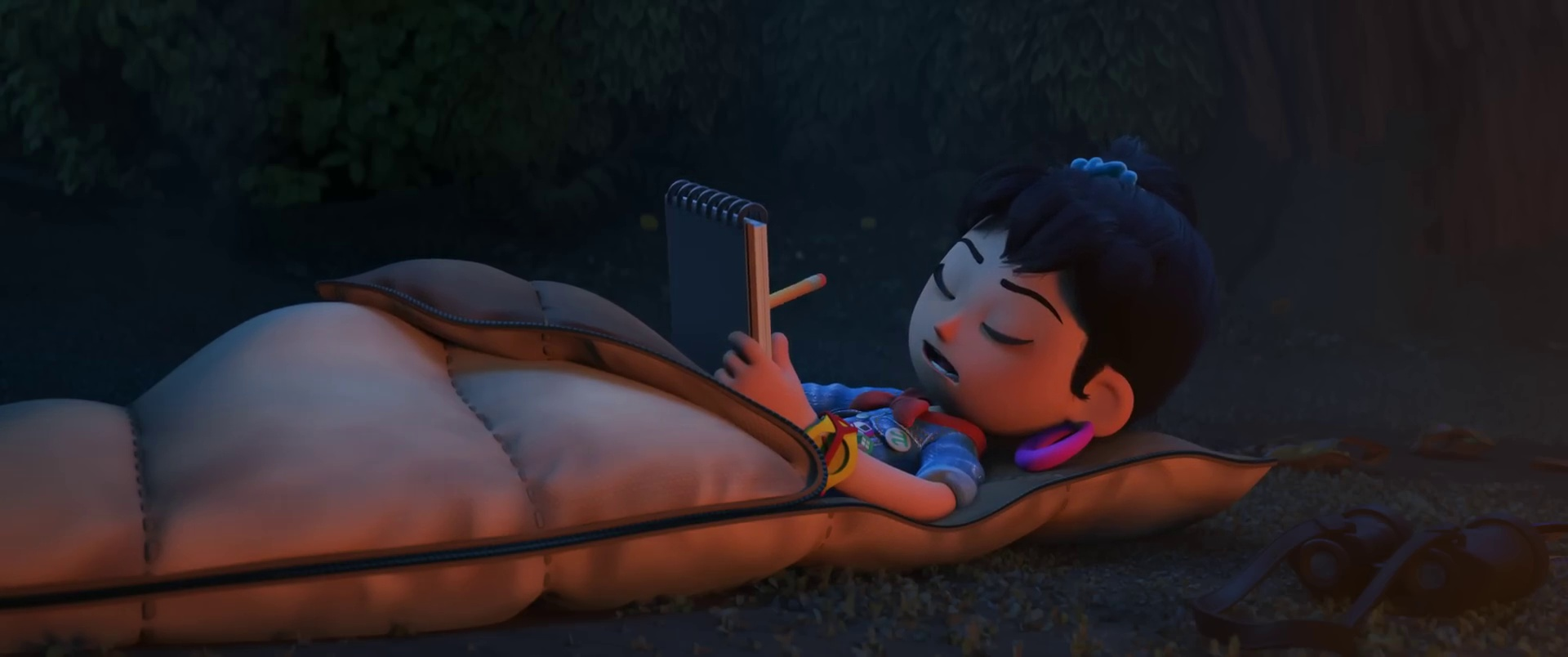}\\
    \texttt{Off-the-shelf captioning pipeline:}
    \begin{lstlisting}[style=caption]
        {
            "scene": "A character is lying in a sleeping bag, holding a notepad and pencil, appearing to be asleep. Another character with vibrant hair is floating in the air with arms outstretched.",
            "background": "The scene is set outdoors at night, with a dark, leafy background and a warm light illuminating the character in the sleeping bag.",
            "characters": {
                "Victoria": {
                    "description": "A character with vibrant, spiky hair, wearing a pink dress with a blue belt, fishnet stockings, and blue shoes.",
                    "location": "Floating in the air in the upper part of the image.",
                    "expression": "Surprised or curious.",
                    "pose": "Arms outstretched, floating."
                },
                "Unknown Character": {
                    "description": "A character with dark hair tied in a bun, wearing a blue shirt with a red scarf and a colorful wristband.",
                    "location": "Lying in a sleeping bag on the ground.",
                    "expression": "Sleeping peacefully.",
                    "pose": "Lying on their back, holding a notepad and pencil."
                }
            },
            "salient_objects": {
                "sleeping bag": "A beige sleeping bag on the ground.",
                "notepad": "A small notepad held by the character in the sleeping bag.",
                "pencil": "A pencil in the character's hand.",
                "binoculars": "A pair of binoculars lying on the ground near the sleeping bag."
            }
        }
    \end{lstlisting}
    \texttt{Improved captioning pipeline:}
    \begin{lstlisting}[style=caption]
        {
            "scene": "Ellie is lying in a sleeping bag, holding a notebook and pencil, appearing to be asleep or resting. She is outdoors at night, surrounded by camping gear.",
            "background": "The scene is set outdoors at night, with dark foliage in the background. The ground is covered with grass and leaves, and there is a warm light illuminating Ellie and her sleeping bag.",
            "characters": {
                "Ellie": {
                    "description": "Ellie has dark hair tied in a ponytail with a blue scrunchie, wearing a denim jacket with a red scarf, khaki shorts, and red boots. She has large pink hoop earrings and a colorful wristband.",
                    "location": "Ellie is lying in a sleeping bag on the ground.",
                    "expression": "Her eyes are closed, and she has a peaceful expression.",
                    "pose": "She is lying on her back, holding a notebook and pencil in her hands."
                }
            },
            "salient_objects": {
                "sleeping bag": "A beige sleeping bag with black stitching, providing warmth and comfort.",
                "notebook": "A small, open notebook held by Ellie.",
                "pencil": "A pencil in Ellie's hand, resting on the notebook.",
                "binoculars": "A pair of binoculars placed on the ground near Ellie's head."
            }
        }
    \end{lstlisting}
    \label{fig:compare_caption_4}
\end{figure}

\end{document}